\documentclass[sn-mathphys,Numbered]{sn-jnl}

\usepackage{graphicx}%
\usepackage{multirow}%
\usepackage{amsmath,amssymb,amsfonts}%
\usepackage{amsthm}%
\usepackage{mathrsfs}%
\usepackage[title]{appendix}%
\usepackage{xcolor}%
\usepackage{textcomp}%
\usepackage{manyfoot}%
\usepackage{booktabs}%
\usepackage{algorithm}%
\usepackage{algorithmicx}%
\usepackage{algpseudocode}%
\usepackage{listings}%

\theoremstyle{thmstyleone}%
\theoremstyle{thmstyletwo}%

\theoremstyle{thmstylethree}%

\begin{document}

\title[Article Title]{Towards the Unification of Generative and Discriminative Visual Foundation Model: A Survey}


\author[1]{\fnm{Xu} \sur{Liu}}

\author[2]{\fnm{Tong} \sur{Zhou}}


\author[3]{\fnm{Yuanxin} \sur{Wang}}

\author[4]{\fnm{Yuping} \sur{Wang}}

\author[5]{\fnm{Qinjingwen} \sur{Cao}}

\author[6]{\fnm{Weizhi} \sur{Du}}

\author[7]{\fnm{Yonghuan} \sur{Yang}}

\author[8]{\fnm{Junjun} \sur{He}}

\author[8]{\fnm{Yu} \sur{Qiao}}

\author[9]{\fnm{Yiqing} \sur{Shen}}\email{yshen92@jhu.edu}
\equalcont{Corresponding Author.}

\affil[1]{\orgdiv{Department of Physics and Astronomy}, \orgname{University of California Los Angeles, Los Angeles}, 
\orgaddress{
\country{USA}}}

\affil[2]{\orgdiv{Department of Computer Science}, \orgname{Rice University}, 
\orgaddress{\country{USA}}}

\affil[3]{\orgdiv{School of Computer Science}, \orgname{Carnegie Mellon University}, 
\orgaddress{\country{USA}}}

\affil[4]{\orgdiv{Department of Electrical and Computer Engineering}, \orgname{University of Michigan}, 
\orgaddress{\country{USA}}}

\affil[5]{\orgdiv{Department of Computer Science}, \orgname{University of Illinois at Urbana-Champaign}, 
\orgaddress{\country{USA}}}

\affil[6]{\orgname{Walmart Global Tech}, 
\orgaddress{\country{USA}}}

\affil[7]{\orgdiv{Department of Computer Science and Engineering}, \orgname{Santa Clara University}, 
\orgaddress{\country{USA}}}

\affil[8]{\orgname{Shanghai AI Laboratory}, 
\orgaddress{\country{China}}}

\affil[9]{\orgdiv{Department of Computer Science}, \orgname{Johns Hopkins University}, 
\orgaddress{\country{USA}}}


\abstract{
The advent of foundation models, which are pre-trained on vast datasets, has ushered in a new era of computer vision, characterized by their robustness and remarkable zero-shot generalization capabilities. 
Mirroring the transformative impact of foundation models like large language models (LLMs) in natural language processing, visual foundation models (VFMs) have become a catalyst for groundbreaking developments in computer vision. 
This review paper delineates the pivotal trajectories of VFMs, emphasizing their scalability and proficiency in generative tasks such as text-to-image synthesis, as well as their adeptness in discriminative tasks including image segmentation. 
While generative and discriminative models have historically charted distinct paths, we undertake a comprehensive examination of the recent strides made by VFMs in both domains, elucidating their origins, seminal breakthroughs, and pivotal methodologies. 
Additionally, we collate and discuss the extensive resources that facilitate the development of VFMs and address the challenges that pave the way for future research endeavors. 
A crucial direction for forthcoming innovation is the amalgamation of generative and discriminative paradigms. 
The nascent application of generative models within discriminative contexts signifies the early stages of this confluence. 
This survey aspires to be a contemporary compendium for scholars and practitioners alike, charting the course of VFMs and illuminating their multifaceted landscape.

}



\maketitle

\section{Introduction}\label{intro}

The advent of foundation models has profoundly revolutionized the field of artificial intelligence (AI). 
These models are distinct in their extensive pre-training on vast datasets, enabling them to exhibit exceptional zero-shot generalization capabilities to unseen data~\cite{bommasani2022opportunities}. 
This ability to adeptly generalize to new datasets and tasks without prior exposure solidifies their position as pivotal elements in the evolving AI landscape.
In the natural language processing (NLP), the impact of foundation models has been striking. 
The development of language foundation models, namely the large language models (LLMs), such as BERT~\cite{bert}, T5~\cite{T5}, and GPTs\footnote{https://cdn.openai.com/papers/gpt-4.pdf} has heralded a paradigm shift, demonstrating an unprecedented level of versatile intelligence. 
These models have successfully unified a myriad of tasks under a single framework, with ChatGPT\footnote{https://openai.com/blog/chatgpt} being a prime example. 
Concretely, powered by GPT-3.5~\cite{InstructGPT} or GPT-4\footnote{https://openai.com/research/gpt-4}, ChatGPT has achieved human-like conversational dynamics, attracting a user base of 173 million and averaging 60 million daily active users as of April 2023.

Simultaneously, the field of computer vision (CV) has witnessed a surge in the exploration of visual models, spurred by the advancements in NLP. 
Accordingly, to enhance the generalization ability to reach the goal of visual foundation models, two primary strategies are employed. 
The first involves increasing the model's size and training it on a larger number of samples. 
For instance, VQ-GAN~\cite{VQ-GAN} integrates approximately 85 million parameters for text-to-image tasks, whereas more sophisticated models like Parti~\cite{parti} encompass around 20 billion parameters. 
The extent of training data also varies markedly; models like StackGAN~\cite{StackGAN} utilize datasets like Oxford-102~\cite{Oxford-120-Flowers} with 8189 examples, contrasting with others like VQ-diffusion that employ the extensive LAION-400M~\cite{LAION-400M} dataset, featuring 400 million image-text pairs.
The second strategy for scaling focuses on augmenting the model's adaptability to a broader array of tasks. 
For example, traditional segmentation networks often had limitations to predefined datasets or tasks~\cite{zhang2021knet}. 
In contrast, contemporary models such as the Segment Anything Model (SAM)~\cite{SAM} have showcased adaptability to various segmentation tasks, both existing and emerging, through prompt engineering, underlining the importance of task generalization.

\begin{figure*}
\centering
\includegraphics[width=1\linewidth]
{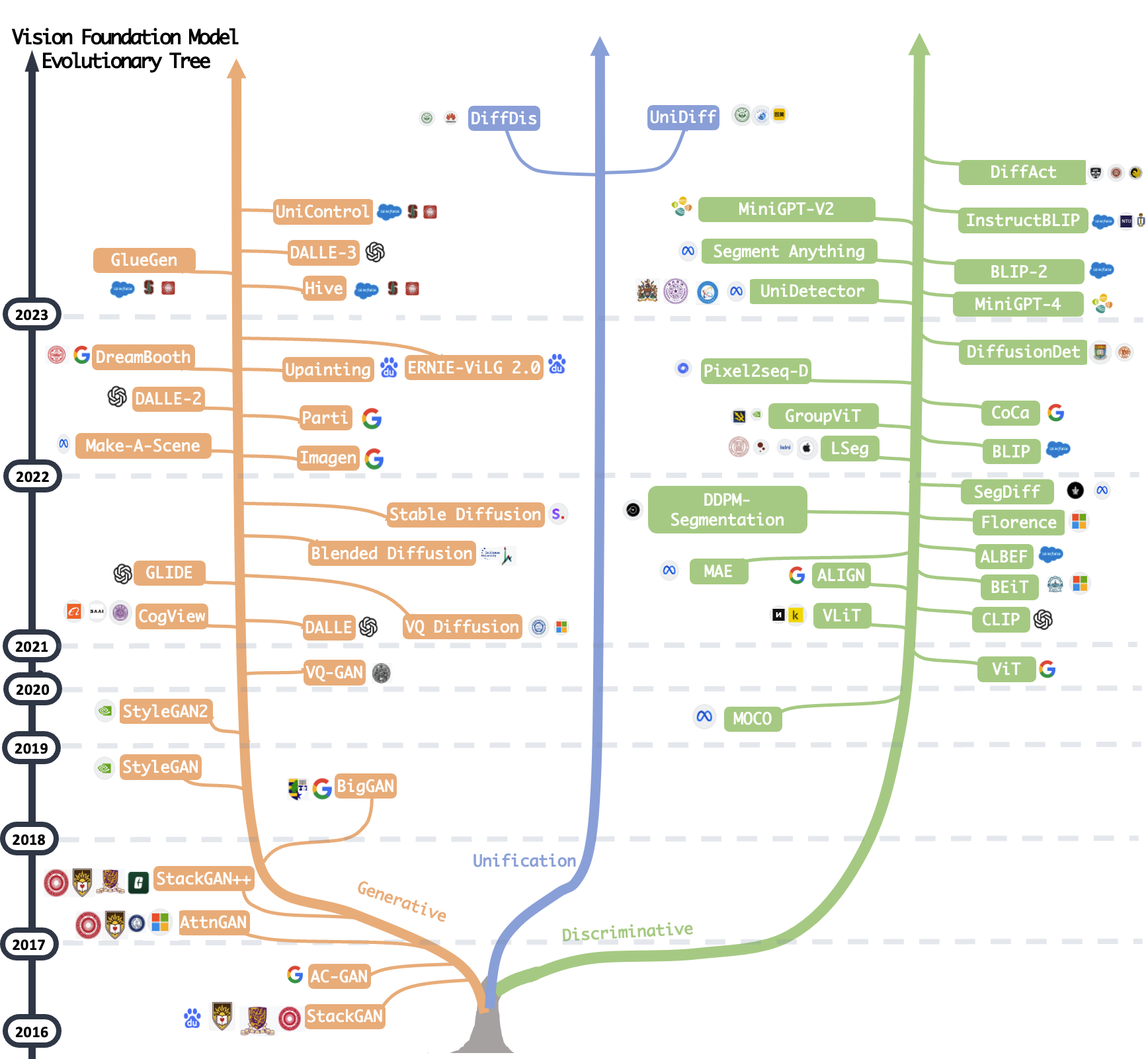}
\caption{A chronological depiction of the evolution of Visual Foundation Models (VFMs). 
The timeline primarily relies on the release dates of the corresponding manuscript, submitted to platforms like arXiv.
In the absence of a paper, the model's earliest public release or announcement date is used as a reference.}
\label{development_history}
\end{figure*}


\subsection{Evolution of Visual Foundation Models}
Figure~\ref{development_history} provides an exhaustive portrayal of the evolution of visual foundation models. 
This illustration separates the progress into two distinct trajectories: discriminative and generative models. 
Traditionally, these trajectories have evolved independently, each harnessing unique techniques to propel their respective domains forward.

In generative modeling, pivotal advancements began with DC-GAN~\cite{DC-GAN}, an early end-to-end differential architecture, extending from characters to pixel levels. Earlier strategies, such as those utilized by StackGAN~\cite{StackGAN} and StyleGAN~\cite{styleGAN}, focused on smaller-scale data. In contrast, later models like DALL-E~\cite{DALLE}, CogView~\cite{cogview}, and Make-A-Scene~\cite{Make-A-Scene} harnessed large-scale datasets. The diffusion model (DM)~\cite{DALLE2,imagen,Glide,stable_diffusion} marks a notable advancement, setting new benchmarks in image synthesis and representing the latest in generative model innovation.
Conversely, in the discriminative domain, exploration has extended to scaling vision transformers, paralleling the emerging capabilities in LLMs. Examples of this progression include ViT-G~\cite{ViT-G}, ViT-22B~\cite{ViT-22B}, Swin Transformer V2~\cite{Swin-Transformer-V2}, and VideoMAE V2~\cite{videomae}. A concurrent effort has been to endow LVMs with multimodal knowledge, as exemplified by models like CLIP~\cite{CLIP} and ALIGN~\cite{ALIGN}, which merge visual and textual data using contrastive learning for zero-shot generalization. Additionally, task-agnostic foundation models like SAM~\cite{SAM} highlight a shift towards a data-centric training approach.

The convergence of generative and discriminative tasks in visual models has a rich history, traceable to early works~\cite{Ng2001OnDV} and energy-based models~\cite{5995710,HINTON2007535}. Many discriminative tasks can effectively be reinterpreted as generative tasks. 
For instance, in segmentation, a generative approach might involve using a prompt like ``cat with black ears'' to generate a precise mask outlining that feature in an image. 
This approach is akin to image inpainting~\cite{li2021semantic} and involves inputs such as text prompts and target images for segmentation.
This synergy demonstrates the capability of generative models to enrich discriminative tasks, with applications in image segmentation~\cite{segdiff,baranchuk2022labelefficient,chen2023generalist, wolleb2021diffusion} and object detection~\cite{chen2023diffusiondet}. 
However, both the academy and industry still face the challenge of developing a unified model that seamlessly combines both generative and discriminative functions in visual foundation models.

\begin{figure}[h!]
\centering
\includegraphics[width=\linewidth]{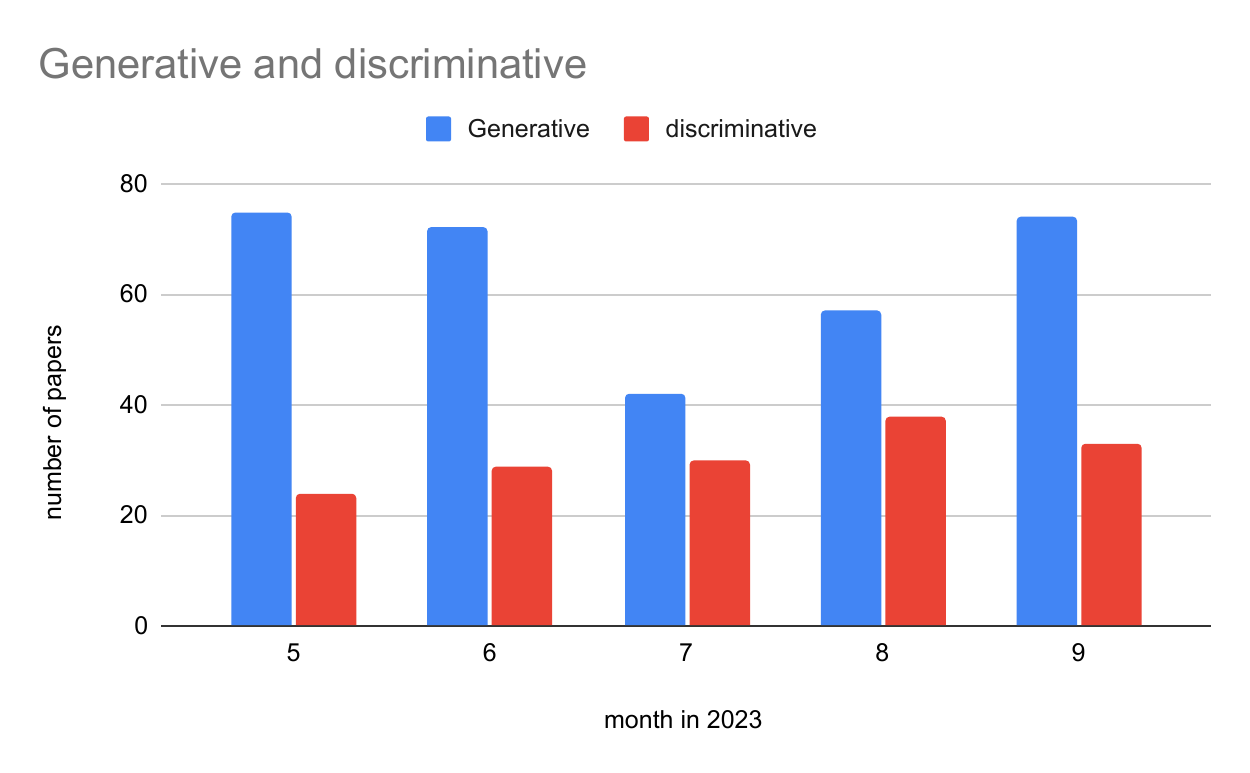}
\caption{This figure illustrates the general trend in the volume of research publications related to visual foundation models. It includes two main categories: 1) Generative tasks, encompassing new models and improvements to existing models, with applications in text-to-image, text-to-video, and text-to-3D generation (relevant search terms on arXiv include "text to image," "video generation," "image editing," "image synthesis," "text to 3D"). 2) Discriminative tasks, covering both foundational and application-oriented research in image classification, segmentation, retrieval, and object detection (search keywords on arXiv: ``vision language model", ``CLIP", ``ALIGN'', ``SAM", ``image classification'', ``image segmentation," ``image retrieval'', ``object detection';). This distribution is crucial for understanding the development and focus areas in the field. }
\label{fig:paper collections}
\end{figure}

\subsection{Comparison to Existing Surveys}
The rapid proliferation of literature on visual foundation models as illustrated in Figure~\ref{fig:paper collections} necessitates a comprehensive and integrated overview, given the current fragmented state of research. 
Existing surveys~\cite{zhou2023vision,zhang2023comprehensive,awais2023foundational}, typically focus on either generative or discriminative paradigms in isolation. 
For instance, the work of Awais et al.~\cite{awais2023foundational} provides a detailed analysis of discriminative models, discussing aspects such as architectural variations, self-supervised learning objectives, large-scale training methodologies, and prompt engineering techniques.
In contrast, this review paper endeavors to bridge the existing gap between these two paradigms. 
We aim to conceptualize a unified perspective that acknowledges the interplay and potential synergies between generative and discriminative tasks in visual foundation models.
Our survey is designed to serve multiple purposes:
\begin{itemize}
    \item Provide a comprehensive reference that encapsulates the latest advancements and methodologies in visual foundation models, offering an inclusive overview of both generative and discriminative paradigms.
    \item Act as a foundational guide for new researchers in the field, presenting a structured and coherent introduction to the key concepts and developments.
    \item Identify and highlight emerging trends and challenges in the field, charting potential future directions and areas of exploration.
    \item Promote the integration and synergy between generative and discriminative tasks, fostering innovation and cross-pollination in research and application.
\end{itemize}

\subsection{Contributions}
\begin{figure*}
\includegraphics[width=\linewidth]{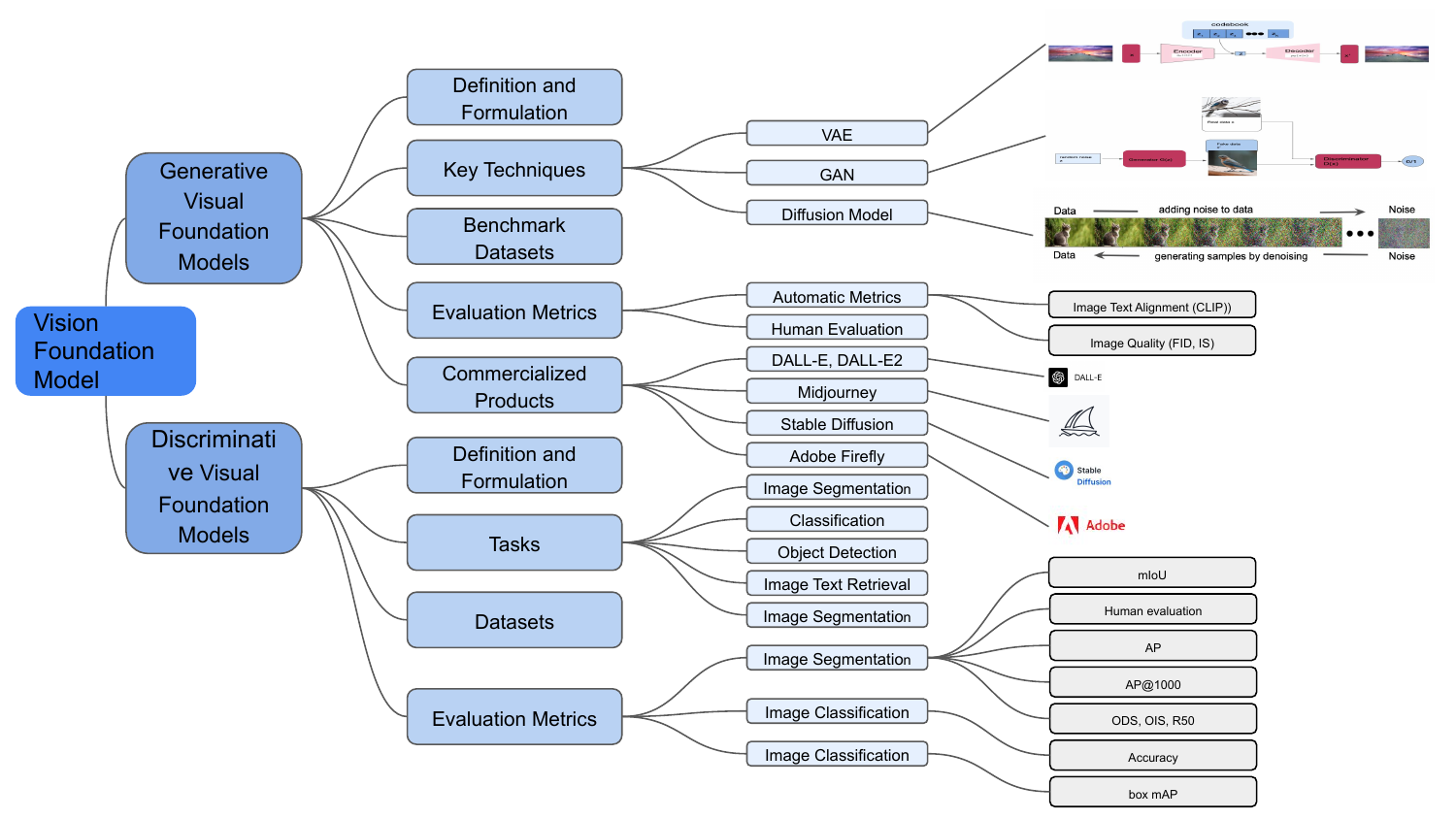}
\caption{The proposed taxonomy of visual foundation models (VFMs). We categorize VFM into generative and discriminative models, depending on the task they focus on.}
\label{fig:taxonomy of visual foundation models}
\end{figure*}

The contributions are summarized as follows.
\begin{itemize}
\item We present a comprehensive taxonomy of visual foundation models, as illustrated in Figure~\ref{fig:taxonomy of visual foundation models}. 
This taxonomy categorizes visual foundation models into two primary groups: Discriminative Visual Foundation Models (DVFM) and Generative Visual Foundation Models (GVFM). 
It aims to provide an exhaustive overview of the field, detailing the distinct functionalities and applications of these models in various computer vision tasks.
\item We critically analyze the differences between generative and discriminative visual foundation models. 
Furthermore, we propose a forward-looking perspective that seeks to integrate these diverse strands of visual foundation models into a cohesive framework.
\end{itemize}

\subsection{Paper Organization}
The existing literature often examines generative and discriminative tasks in isolation, leaving a gap for a unified analysis. 
This survey aims to fill this void by presenting a holistic view of the recent advancements of foundation models in the image domain, with a particular focus on examining the foundation model from discriminative and generative perspectives. 
The paper is structured as follows:
\begin{itemize}
\item Section 2: Provides an in-depth examination of foundational models, contrasting large language models with their visual counterparts.
\item Section 3: Focuses on the technological foundations and diverse applications of generative visual foundation models.
\item Section 4: Discusses the architectural nuances of discriminative visual foundation models and their various implementations.
\item Section 5: Addresses multimodal visual foundation models, exploring their integration and interplay between different modalities.
\item Section 6: Investigates the limitations of current visual foundation models and outlines potential avenues for future research.
\item Section 7: Concludes the paper, summarizing key insights and takeaways from the survey.
\end{itemize}

\section{Foundation Model}

\subsection{Definition of Foundation Model}
Historically, deep learning models have predominantly been anchored in supervised learning paradigms. 
A notable limitation of these methods is their substantial reliance on extensive manual annotations, which are often costly and time-consuming to obtain~\cite{bommasani2022opportunities}. 
This dependence restricts the models' capability to generalize effectively across varying scenarios and limits their broader application in diverse fields.

In response to these constraints, \textit{foundation models} have emerged as a transformative approach, shifting away from the heavy dependence on labeled data. 
Characterized by their use of self-supervised learning pre-training, these models leverage an extensive array of datasets. 
Consequently, this approach allows them to operate beyond the confines of specific tasks~\cite{bomma}.
That is to say, foundation models are renowned for their adaptability and versatility. 
They can be fine-tuned to a multitude of downstream tasks, attaining proficiency in new and specific areas through additional task-specific training~\cite{bomma}. 
This adaptability is a defining trait of foundation models, which is exemplified in two distinct categories: \textit{Language Foundation Models} (LFMs) and \textit{Visual Foundation Models} (VFMs).

Language foundation models, such as GPT-2~\cite{GPT-2}and GPT-3~\cite{GPT-3}, have made significant strides in the AI field, showcasing an exceptional understanding and generation of human language. 
Building on this momentum, the focus has increasingly shifted towards visual foundation models. 
Models like SAM~\cite{SAM} and DALL-E2~\cite{DALLE2} are prime examples in this category, underscoring the potential of extending foundational principles to visual computing.


\subsection{Language Foundation Models}
Language models are designed to predict and generate sequences of words, capturing the underlying structure of language. 
The evolution of language foundation models can be segmented into distinct phases, each marked by significant advancements.

\paragraph{Pretrained Language Models}
Early language models, utilizing neural networks such as Recurrent Neural Networks (RNNs), were primarily focused on estimating the likelihood of word sequences~\cite{kombrink2011recurrent, mikolov2010recurrent}. 
The introduction of \textit{word2vec} represented a pivotal shift, simplifying the neural network approach to learning effective word representations, thereby enhancing performance across various NLP tasks~\cite{mikolov2013efficient}.
This marked the inception of representation learning in language models, extending their utility beyond mere word sequence modeling.
\textit{Pretrained Language Models} (PLMs) revolutionized this domain by learning universal language representations, beneficial for a wide range of downstream NLP tasks, thereby obviating the need to train models from scratch for each new task~\cite{liu2023pre}.
The renaissance of PLMs was significantly propelled by the advent of the Transformer architecture, which brought the self-attention mechanism to the forefront. 
BERT~\cite{bert} exemplified this architectural evolution by pretraining bidirectional language models on extensive unlabeled text, achieving unprecedented performance in context-rich word representation across diverse NLP tasks. 
This milestone not only spurred intensive research but also established the `pre-training and fine-tuning' approach as a fundamental methodology for modern LFMs. 
This period saw the emergence of various PLMs, such as GPT-2~\cite{GPT-2} and BART~\cite{BART}, each exhibiting unique architectural innovations or refining pretraining techniques~\cite{RoBERTa, sanh2022multitask, wang2022language}. 
Fine-tuning remains essential in tailoring these broad-spectrum models for specific task requirements.

\paragraph{Large Language Models} 
The evolution of PLMs into larger-sized models, known as \textit{Large Language Models} (LLMs), marked a significant leap in their capabilities. 
Scaling laws, involving increases in both model size and data volume, have been crucial in this evolution~\cite{kaplan2020scaling}. 
This scaling has consistently led to improved performance across various downstream tasks. 
Models like the 175-billion parameter GPT-3~\cite{GPT-3} and the 540-billion parameter PaLM are testaments to the enhanced capabilities achieved through scaling.
Moreover, LLMs have displayed emergent abilities not observed in their smaller counterparts, such as BERT~\cite{bert} or GPT-2~\cite{GPT-2}. 
For instance, GPT-3 demonstrates a remarkable proficiency in few-shot learning through in-context adaptation, a skill-less pronounced in smaller models. 
A practical application of LLM's abilities is evident in systems like ChatGPT, which leverages the advanced architecture of LLMs to deliver nuanced conversational interactions.


\paragraph{Taxonomy for LFM}
Previous surveys~\cite{zhao2023survey} on LFMs broadly categorize them based on their operational tasks. 
These tasks are typically divided into generative and discriminative categories. 
Generative tasks include activities such as language generation~\cite{li2021pretrained,bahdanau2016neural,rush2015neural,chen2017reading} and complex reasoning~\cite{zhou2023leasttomost}, where the model generates new text based on input. 
Discriminative tasks, such as text classification\cite{raffel2023exploring}, involve categorizing or interpreting given texts.

LFMs are also classified according to their architectural designs, primarily falling into three distinct categories~\cite{amatriain2023transformer}:
\begin{itemize}
    \item \textit{Encoder-only Models:} An exemplar of this category is BERT~\cite{bert}, which employs an encoder-only architecture. BERT, with approximately 300 million parameters, utilizes pretraining followed by a fine-tuning paradigm. 
    It focuses on masked language models as the core training objective during pretraining, subsequently adapting the pre-trained model for annotated downstream datasets. 
    Models in this category are particularly suited for discriminative tasks due to their robust feature extraction capabilities.
    
    \item \textit{Decoder-only Models:} GPT~\cite{GPT-2} and its successors epitomize decoder-only models. Utilizing the decoder component of an auto-regressive transformer model, GPT models, including GPT-3~\cite{GPT-3} with 175 billion parameters, are adept at predicting the next token in a sequence. 
    GPT models also adhere to the pretraining and fine-tuning paradigm, with GPT-3 introducing an innovative approach by framing all NLP tasks as generating textual responses based on given prompts\cite{GPT-3}. 
    This makes them highly effective for generative tasks, where the creation of coherent and contextually relevant text is paramount.

    \item \textit{Encoder-Decoder Models:} T5~\cite{T5}, with an 11 billion parameter encoder-decoder transformer model, represents this category. 
    Such models are designed to generate new sentences based on given inputs, following the standard pretraining and fine-tuning paradigm\cite{soltan2022alexatm}. 
    The encoder-decoder architecture equips these models with the flexibility to handle both generative and discriminative tasks, making them versatile tools in various NLP applications\cite{amatriain2023transformer}.
\end{itemize}
This taxonomy delineates the operational and architectural distinctions among different LFMs, providing clarity on their specific functionalities and suitable applications in the domain of natural language processing.

\subsection{Visual Foundation Models}
\textit{Visual Foundation Models} (VFMs) are rapidly gaining prominence in the field of computer vision (CV), mirroring the transformative impact of LFM in the NLP domain. 
These models have proven their versatility, excelling in a range of tasks from generating realistic images and videos~\cite{DALLE2, stable_diffusion} to performing image classification~\cite{CLIP}, image segmentation~\cite{SAM}, and object detection~\cite{unidetector}.

\paragraph{Pretrained Vision Models}
In CV, foundational models often draw from the advancements in LLMs, showcasing a fusion of principles and architectures. 
Notable examples of this synergy include CLIP~\cite{CLIP}, ALIGN~\cite{ALIGN}, Florence~\cite{Florence}, VLBERT~\cite{vlbert}, and X-LXMERT~\cite{xlxmert}. 
These models are trained on extensive datasets to produce text-image paired embeddings. 
These embeddings are then leveraged in various specialized models designed for specific visual tasks. 
Echoing the approach of LLMs, these  \textit{Pretrained Vision Models} are capable of learning universal visual representations, significantly benefiting a wide array of downstream CV tasks.

\paragraph{Our Taxonomy for Visual Foundation Models}
VFMs, akin to their language counterparts, typically address two primary task categories: generative and discriminative. 
Generative models in VFMs focus on understanding and replicating the underlying data distribution of visual elements, such as images and videos. 
This understanding enables them to synthesize new, realistic visual content. 
On the other hand, discriminative models are tailored to establish clear decision boundaries among various categories or classes, leading to enhanced performance in tasks like image classification and segmentation.
However, unlike language models, there is currently no unified model in the visual domain that can adeptly handle both task types despite the distinct strengths of each task type.
Bridging this divide presents a significant challenge and opportunity in the evolution of VFM research.


\section{Generative Visual Foundation Models}

\subsection{Definition and Formulation}

\paragraph{Generative Visual Model}
Generative models aim to model and replicate the inherent probability distribution present within a dataset. 
Specifically, \textit{Generative Visual Models} (GVMs) are generative models that focus on visual data such as images and videos. 
These models strive to emulate the probability distribution $p(\mathbf{x})$ associated with visual data $\mathbf{x}$. 
This emulation is achieved through a process of parameterization, expressed as $p_\theta(\mathbf{x})$, where $\theta$ represents the learnable parameters of the model. 
The effectiveness of GVMs is evident in their ability to generate novel data samples that are reminiscent of the original dataset. 
Recent advancements have significantly enhanced the capabilities of GVMs, solidifying their role in the domain of AI-generated content (AIGC). 
This encompasses the generation of high-quality images, engaging videos, and detailed image-to-image translations, as showcased by various models such as conditional GAN, DALL-E, Imagen, Stable Diffusion and \textit{etc}~\cite{ConditionalGAN,DALLE,imagen,stable_diffusion,makeavideo,imagen_video,tumanyan2022plugandplay,isola2018imagetoimage,Richardson_2021_CVPR}. 

\paragraph{Generative Visual Foundation Models}
While conventional GVMs excel in various aspects of visual content generation, they often encounter task-specific limitations. 
For example, Generative Adversarial Networks (GANs), a subset of GVMs, grapple with challenges like model collapse and a lack of diversity~\cite{VQ-VAE-2,DiffusionBeatGANs}. 
These issues hinder their scalability and adaptability to new domains. 
Their constrained flexibility and heavy dependence on specific training datasets limit their broader applicability.
In response, \textit{Generative Visual Foundation Models} (GVFMs) have been developed to address the shortcomings of traditional GVMs. 
GVFMs are characterized as advanced GVMs trained on extensive and varied datasets, enabling them to adapt flexibly, often through fine-tuning, to a wide array of downstream tasks~\cite{DALLE2}. 
These tasks range from text-to-image generation to inpainting~\cite{lugmayr2022repaint}, super-resolution~\cite{li2021srdiff,saharia2021image}, and image editing~\cite{chandramouli2022ldedit,couairon2023diffedit}.
Specifically, GVFMs utilize large-scale datasets and self-supervision techniques to build robust and versatile data representations~\cite{bommasani2022opportunities}. 
Their architectural design is modular, facilitating easy adaptation for diverse applications across numerous fields. 
GVFMs stand out in their ability to tackle complex visual content generation challenges, extending their capabilities beyond mere visual synthesis to include altering existing visuals and creating content from textual prompts. 
Leading models in this category, such as LDM~\cite{stable_diffusion}, DALL-E~\cite{DALLE}, DALL-E 2~\cite{DALLE2}, Imagen~\cite{imagen}, and GLIDE~\cite{Glide}, exemplify the extensive potential and transformative impact of GVFMs in the arena of AI-generated content. A summary of these generative visual foundation models can be found in Table~\ref{tab:generative_models}.

\subsection{Key Techniques in Generative Visual Foundation Models}
GVFMs are grounded in three principal techniques that constitute their method foundation, namely (1) Variational Autoencoders (VAEs), (2) Generative Adversarial Networks (GANs), and (3) Diffusion Models.

\paragraph{VAE}
VAEs are an advanced form of the traditional autoencoder architecture. 
Autoencoders typically compress input data into a lower-dimensional latent space through an encoder and then reconstruct the input from this latent representation via a decoder~\cite{kingma2022autoencoding}. 
VAEs refine this approach by modeling the data distribution $p(\mathbf{x})$ using a latent space feature $\mathbf{z}$, formulated as $p(\mathbf{x}) = \int p(\mathbf{x}|\mathbf{z})dp(\mathbf{z})$. 
In this framework, the decoder estimates $p(\mathbf{x}|\mathbf{z})$, while the encoder approximates the posterior distribution $p(\mathbf{z}|\mathbf{x})$ using Bayes' theorem. 
VAEs introduce a probabilistic component to the autoencoding process, enabling them to generate a diverse array of samples from the latent space, thus enhancing the model's capability to explore and interpolate within the data space.

\paragraph{GAN}
A GAN comprises two interconnected neural network models: a generator $G$ and a discriminator $D$~\cite{GAN}. 
These two models engage in a strategic game, where the generator aims to produce data samples that resemble real data, and the discriminator tries to differentiate between real and generated samples. 
The interaction is governed by the following objective:
\begin{equation}
\min_G \max_D E_{x\sim p(x)}[\log D(x)] +E_{z\sim p_z(z)}[\log (1-D(G(z)))].
\end{equation}
In this setup, the generator $G$ is trained to create convincing data samples, while the discriminator $D$ learns to accurately identify real versus generated samples.

\paragraph{Diffusion Model}
Denoising Diffusion Probabilistic Models (DDPMs) utilize a pair of Markov chains to transform data into noise and then revert it back to its original form~\cite{DPM,DDPM,song2022denoising, song2021scorebased,jolicoeurmartineau2020adversarial}. 
The forward chain incrementally adds noise to the data through a series of steps $\mathbf{x}_1, \mathbf{x}_2, ... \mathbf{x}_T$, following a transition kernel $q(\mathbf{x}_t|\mathbf{x}_{t-1})$. 
The joint distribution of this process is expressed as:
\begin{equation}
q(\mathbf{x}_1, ..., \mathbf{x}_T|x_0)=\prod_{t=1}^T q(\mathbf{x}_t|\mathbf{x}_{t-1}),
\end{equation}
with Gaussian perturbation as the transition kernel:
\begin{equation}
q(\mathbf{x}_t|\mathbf{x}_{t-1})=N(\mathbf{x}_t;\sqrt{1-\beta_t}\mathbf{x}_{t-1},\beta_t),
\end{equation}
where $\beta_t$ is a predefined hyperparameter.
The reverse Markov chain begins with a prior distribution $p(\mathbf{x}_T) = N (\mathbf{x}_T; 0, \mathbf{I})$ and employs a trainable transition kernel $p_\theta(\mathbf{x}_{t-1}|\mathbf{x}_t)$, which is characterized as:
\begin{equation}
   p_\theta(\mathbf{x}_{t-1}|x_t) = N (\mathbf{x}_{t-1};\mu_\theta(\mathbf{x}_t,t), \Sigma_\theta(\mathbf{x}_t,t)), 
\end{equation}
where $\theta$ denotes the model parameters. 
Data reconstruction occurs by starting with a noise vector $\mathbf{x}_T$ drawn from $p(\mathbf{x}_T)$ and iteratively applying the transition kernel until reaching $t=1$. 

\subsection{Autoregressive GVFMs}
Autoregressive Encoder (AE) models have shown significant promise in generating data sequences, applicable to both text and visuals. 
These models function by predicting subsequent tokens based on their predecessors, ensuring the output is contextually relevant. 
However, the sequential nature of autoregressive models often results in high computational demands and can accumulate errors over extended sequences, posing challenges for scalability and accuracy.

\paragraph{Cogview and Cogview2}
Cogview and its enhanced version, Cogview2~\cite{cogview}, represent significant strides in autoregressive GVFMs. 
These models employ large-scale joint pretraining to create both textual and visual tokens. 
The visual tokens are extracted using Vector Quantized Variational AutoEncoder (VQ-VAE)~\cite{VQ-VAE}, a method that enables the effective encoding of visual information into a compressed format, later used for generating complex visual outputs.

\paragraph{DALL-E}
DALL-E~\cite{DALLE} introduces an innovative approach in autoregressive GVFMs. 
It is a two-stage model that utilizes a transformer to autoregressively process both text and image tokens as a single, unified data stream. 
This integration allows for a more cohesive and contextually aligned generation of visual content from textual descriptions.

\paragraph{Parti}
Pathways Autoregressive Text-to-Image (Parti)~\cite{parti} treats text-to-image generation as a sequence-to-sequence task, akin to machine translation. 
This model operates in two phases: initially, an image tokenizer converts images into a sequence of visual tokens. 
In the second phase, an autoregressive model is employed to generate these image tokens based on the provided text tokens, effectively bridging the gap between textual input and visual output.

\paragraph{Make-a-scene}
Make-a-scene~\cite{Make-A-Scene} takes autoregressive modeling further by incorporating implicit conditioning with controlled scene tokens derived from segmentation maps. 
This method allows for more precise and contextually appropriate visual content generation, as it extends the capability of the model beyond traditional text and image tokens, offering enhanced control over the generated visual scenes.

\subsection{VAE-based GVFMs}
VAEs have evolved into powerful generative models, playing a pivotal role in the development of GVFM. 
This section explores prominent VAE architectures and their integration into visual foundation models, highlighting their unique contributions and synergies.

\paragraph{Vector Quantized-Variational Autoencoder (VQ-VAE)}
The VQ-VAE framework is a notable advancement in VAE, introducing discrete latent variables to train an encoder. 
This approach effectively compresses images into a discrete, low-dimensional latent space~\cite{VQ-VAE}. 
A key advantage of VQ-VAE is its resolution of the ``posterior collapse'' issue, a common challenge in VAE architectures with powerful decoders where latent variables lose efficacy~\cite{lucas2019understanding}. 
VQ-VAE also addresses variance challenges. 
While not a visual foundation model in the strictest sense, its discrete latent space has been instrumental in models like VQ-Diffusion~\cite{VQ-diffusion}. 
The fusion of VQ-VAE with diffusion models showcases how discrete latent variables can enhance image quality and enable a more nuanced representation of variable dependencies.

\paragraph{Hierarchical Variational Autoencoder}
Hierarchical Variational Autoencoders (HVAEs) represent an extension of the vanilla VAE, introducing multiple layers of hierarchy in latent variables. 
In this architecture, latent variables are derived from higher-level, more abstract latent. 
The Very Deep VAE (VD-VAE)~\cite{VD-VAE} is a leading example, outperforming autoregressive models like PixelCNN on natural image benchmarks in terms of log-likelihood. 
The VQ-VAE2~\cite{VQ-VAE-2} is another noteworthy model, combining a two-tier hierarchical VQ-VAE with an autoregressive PixelCNN as its prior. 
This model utilizes hierarchical multi-scale latent maps to enhance resolution while maintaining the efficient encoder-decoder architecture of the original VQ-VAE.

\paragraph{VAEs as Visual Foundation Models}
Although VAEs and flow-based models are proficient in generating high-resolution images, the visual quality they achieve may not always match that produced by GANs. 
However, their incorporation into diffusion models presents an intriguing avenue for advancement. 
For example, DiffuseVAE~\cite{pandey2022diffusevae} integrates a traditional VAE within the DDPM. 
It conditions the diffusion sampling process using VAE-generated blurry image reconstructions. 
This highlights how VAEs, when combined with other generative approaches, can enhance the robustness and adaptability of visual foundation models, contributing significantly to their overall performance and versatility.

\subsection{GAN-based GVFMs}
\label{section:GAN}
GANs have evolved as a key counterpart to VAE models, especially in generative tasks that integrate text and image synthesis, such as text-to-image generation.

\paragraph{Conditional GANs and Variants}
Conditional generation has become a cornerstone in GVFMs, with GANs leading the charge in generating images from textual descriptions. 
The conditional GAN (cGAN)~\cite{ConditionalGAN}, directed by class labels, exemplifies this concept. 
It focuses on producing images that align visually with specified classes. 
An evolution of this model is the AC-GAN (auxiliary classifier GAN)~\cite{AC-GAN}, which integrates accurate class label classification into the generative process. 
This approach not only enhances the visual appeal of the generated samples but also ensures they are categorically accurate, highlighting the importance of class information in image synthesis.

\paragraph{DC-GAN}
Deep Convolutional GAN (DC-GAN)~\cite{DC-GAN} represents a significant leap in GAN development by conditioning the generative process on textual descriptions rather than class labels. 
This innovative approach facilitates a direct, end-to-end differentiable transformation from textual sequences to pixel arrays, opening up new possibilities for image synthesis driven by text.

\begin{figure}[h!]
 \centering
\includegraphics[width=\linewidth]
{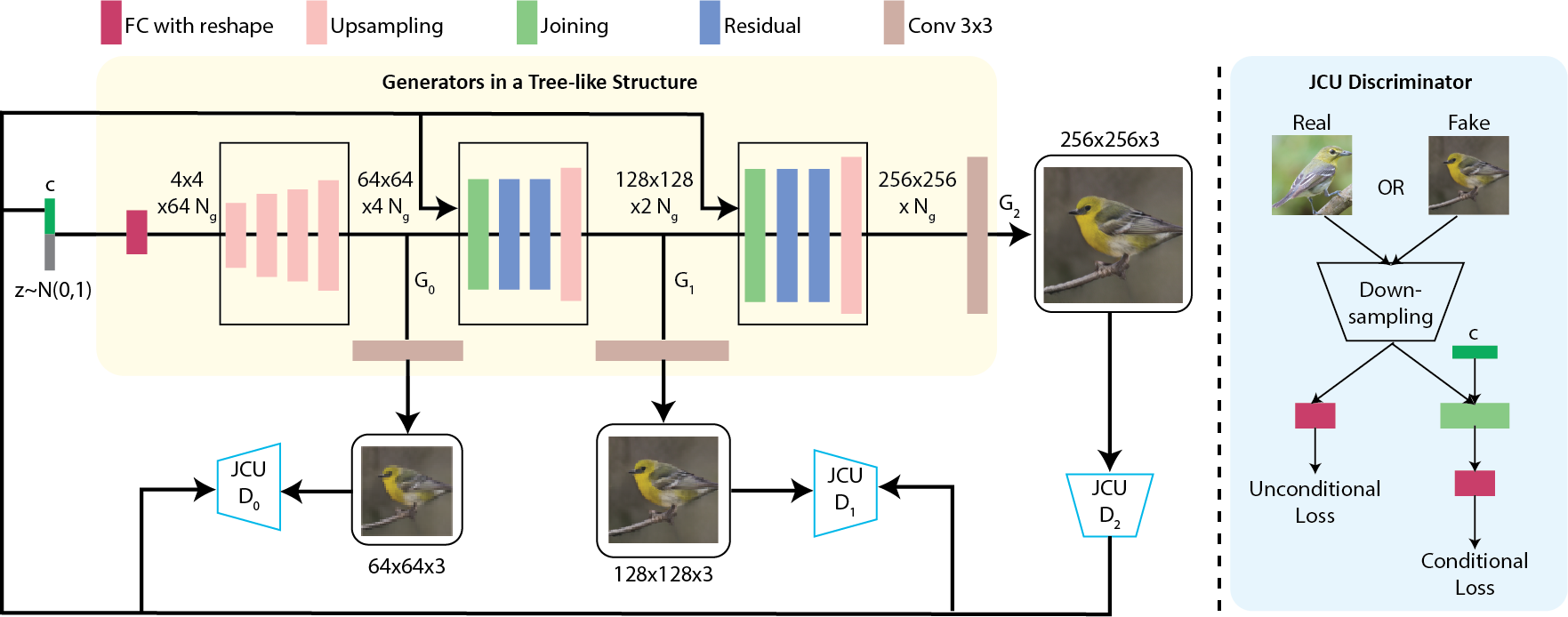}
\caption{A graphical representation of the StackGAN architecture, showcasing its innovative tiered approach for generating high-resolution images from descriptive text.}
\label{fig:StackGAN}
\end{figure}

\paragraph{Stacked-Structure GANs and Variants}
While cGAN and DC-GAN have transformed text-related image generation, producing high-resolution, photo-realistic images remains computationally challenging. 
Stacked-structure GANs are designed to overcome these obstacles. 
As illustrated in Fig.~\ref{fig:StackGAN}, StackGAN~\cite{StackGAN} utilizes a two-stage generation process. 
The Stage-I GAN forms basic shapes and colors from text, generating low-resolution images. 
The Stage-II GAN then enhances these images, infusing them with high-resolution details. 
This progressive approach allows for the creation of more refined and detailed visual content.
Building on StackGAN, StackGAN++~\cite{StackGAN++} implements a tree-structured architecture with multiple generators and discriminators at different scales. 
This approach allows for the creation of images at varying resolutions while maintaining consistency within the scene. 
It improves training stability and mitigates overfitting, as demonstrated by similar architectures like HDGAN and other high-resolution GANs~\cite{HDGAN, wang2018highresolution}.

\paragraph{AttnGAN}
AttnGAN~\cite{xu2017attngan} advances the capabilities of GANs in text-to-image translation by implementing a multi-stage refinement strategy combined with an attention mechanism. 
This model addresses the limitations of stacked-structure GANs in generating fine-grained details, thereby enhancing the precision and quality of the synthesized images.

\paragraph{StyleGAN and Evolution}
StyleGAN~\cite{styleGAN} leverages a style-based generator, drawing inspiration from early style transfer techniques. 
Its successor, StyleGAN2~\cite{styleGAN2}, introduces generator regularization and other improvements, significantly enhancing image quality and fidelity.

\begin{figure}[h!]
 \centering
\includegraphics[width=\linewidth]
{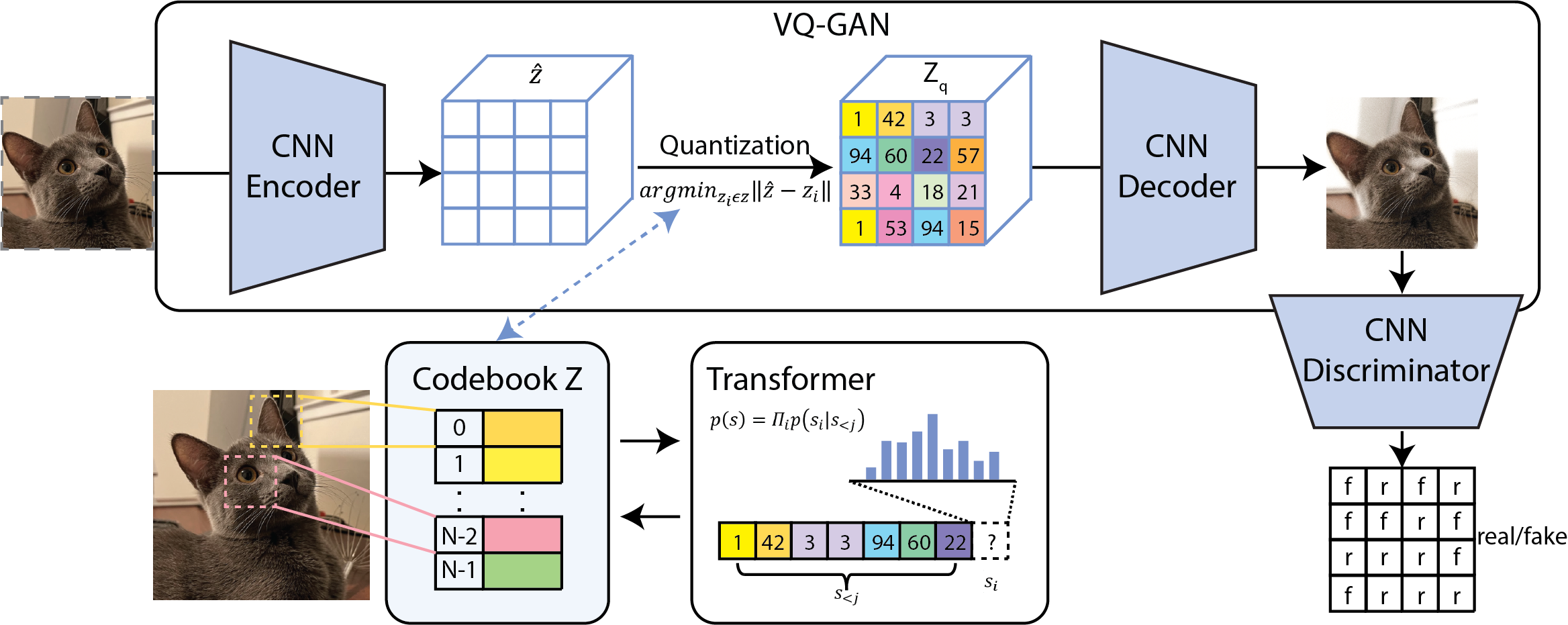}
\caption{The architecture of VQ-GAN demonstrating the integration of CNN and Transformer models for image reconstruction and generation.}
\label{fig:VQ-GAN}
\end{figure}

\paragraph{VQ-GAN}
VQ-GAN~\cite{VQ-GAN} integrates the strength of CNNs with the adaptability of transformers to produce high-definition images, as depicted in Figure~\ref{fig:VQ-GAN}. 
This innovative model harmonizes the distinct advantages of both architectures to create a cohesive and effective image generation framework.

\paragraph{BigGAN}
BigGAN~\cite{BigGAN} builds upon the principles of SAGAN, underscoring the impact of scaling up GAN training by increasing layer channel counts and batch sizes. 
This approach has proven effective in boosting model performance and enhancing image quality.

\paragraph{Emerging GAN Variants}
Innovative GAN variants such as DM-GAN~\cite{zhu2019dmgan}, Object-GAN~\cite{objectGAN}, and ControlGAN~\cite{ControlGAN} have emerged, each offering unique improvements to the field of GANs.

\paragraph{GANs as Visual Foundation Models}
GANs have become fundamental in creating high-resolution and perceptually authentic images, despite challenges in optimization and fully capturing data distributions. 
Their versatility and effectiveness in various image generation tasks underscore their central role in the visual foundation model landscape.

\subsection{Diffusion Model based GVFMs}

Diffusion models have emerged as significant players in the realm of visual foundation models, acclaimed for their stationary training objectives and exceptional scalability. These models have been increasingly recognized for their efficacy in vision-related tasks.

\paragraph{ADM and ADM-G}
The Architectural Diffusion Model (ADM)~\cite{DiffusionBeatGANs} marks a milestone in the diffusion model landscape, surpassing the performance of GANs through refined model architecture and a strategic balance between diversity and fidelity. Building upon ADM, ADM-G introduces Classifier guidance, a concept previously discussed in Section~\ref{section:GAN}. This enhancement enables ADM-G to further elevate the quality of diffusion models, enriching their application in complex vision tasks.

\paragraph{GLIDE and Classifier-Free Guidance}
While classifier guidance boosts the performance of models like ADM-G, it also introduces additional complexities and potential biases~\cite{ho2022classifierfree}. Classifier-free guidance was developed to mitigate these challenges, streamlining the diffusion model training process. GLIDE~\cite{Glide} effectively utilizes this approach in text-to-image synthesis, replacing class labels with textual descriptions. Despite initial explorations with CLIP guidance, GLIDE's evaluations favored classifier-free guidance for its authentic outputs that closely align with the provided captions.

\paragraph{Imagen}
Taking inspiration from GLIDE, Imagen~\cite{imagen} integrates classifier-free guidance into its image synthesis framework. In contrast to GLIDE's simultaneous training of the text encoder, Imagen leverages a pre-trained, static large language model, tapping into the profound textual understanding capabilities of advanced transformer models.

\paragraph{Latent Diffusion Models}
Direct image generation from high-dimensional pixels, as in GLIDE and Imagen, entails substantial computational requirements. To address this, some models have adopted a strategy of compressing images into a low-dimensional latent space before processing. Notable models employing this strategy include Latent Diffusion Models (LDMs) like VQ-diffusion and DALL-E 2.

\paragraph{VQ-Diffusion}
Vector Quantized Diffusion (VQ-Diffusion)~\cite{VQ-diffusion} extends the VQ-VAE framework by integrating its latent space with a conditional version of the DDPM. This combination offers a nuanced approach to image generation, capitalizing on the strengths of both diffusion and autoencoder technologies.

\begin{figure}[h!]
 \centering
\includegraphics[width=\linewidth]
{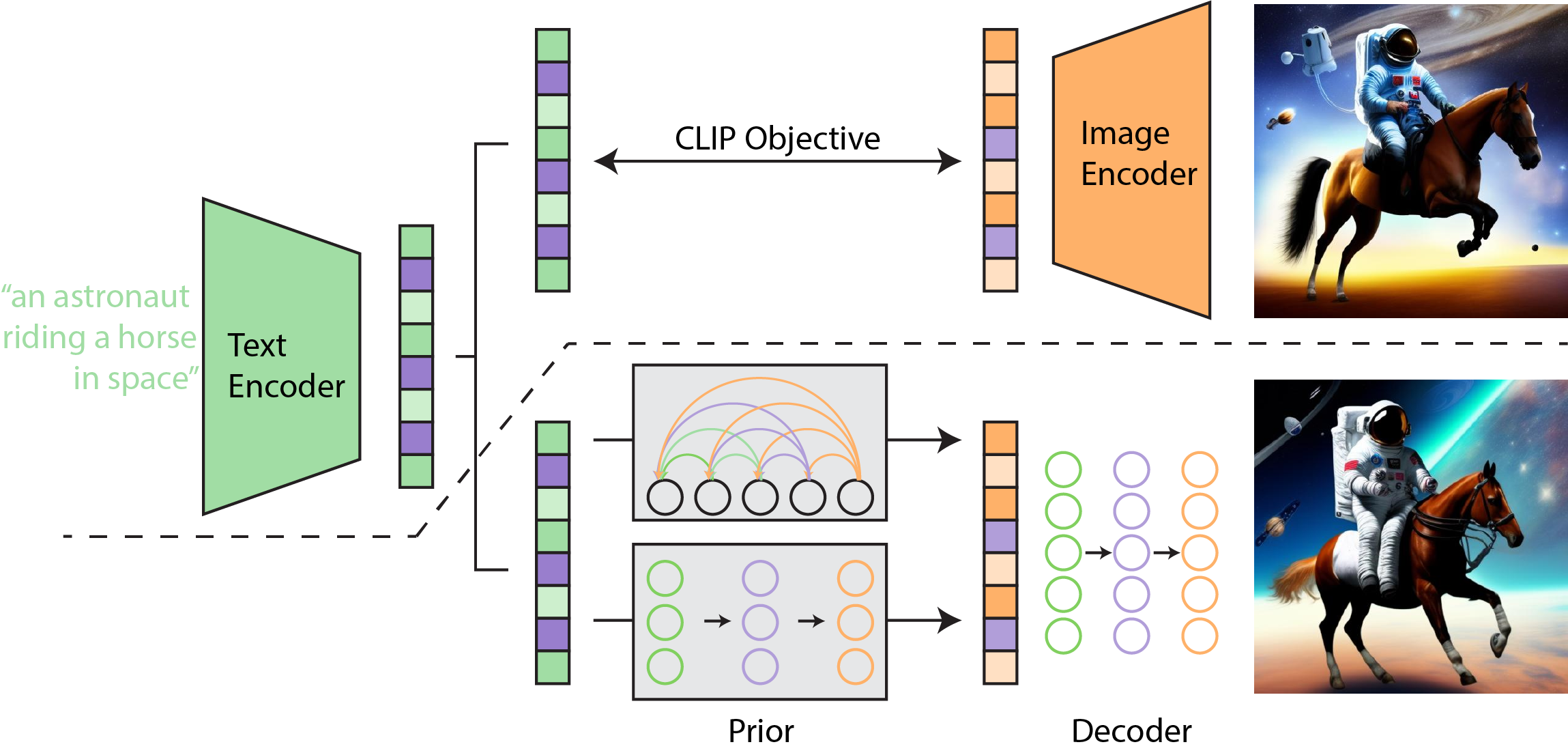}
\caption{DALL-E 2's innovative integration of text and image encoders to synthesize images from descriptive text.}
\label{fig:Dalle2}
\end{figure}

\paragraph{DALL-E 2}
Illustrated in Figure~\ref{fig:Dalle2}, DALL-E 2~\cite{DALLE2} represents a confluence of CLIP's embedding techniques and diffusion model principles. It commences by generating a CLIP image embedding from text, which is then used by a decoder to produce an image. This tiered system harnesses the power of both CLIP's multimodal embedding and the generative capabilities of diffusion models, making DALL-E 2 a formidable tool in visual content creation.

\paragraph{Upainting}
Upainting~\cite{Upainting} represents a significant advancement in the field of text-to-image synthesis, combining architectural innovations with diverse guidance strategies. 
It effectively incorporates cross-modal guidance from a pre-trained image-text matching model into a text-conditional diffusion model. 
The integration of a pre-trained Transformer language model as the text encoder allows Upainting to leverage the comprehensive language understanding capabilities of large-scale Transformer models. 
Combined with image-text matching models, this approach effectively assimilates cross-modal semantics and styles. 
The result is a notable enhancement in sample fidelity, ensuring that the generated images are more closely aligned with the textual descriptions, thereby bridging the gap between language and visual representation.

\paragraph{Blended Diffusion}
Blended Diffusion~\cite{Avrahami_2022} harnesses the strengths of pre-trained DDPM~\cite{DDPM} and CLIP~\cite{CLIP} to offer a novel solution for region-specific image editing. 
This model applies natural language instructions to guide the editing process, accommodating a diverse range of real images. 
Its universal applicability and ability to facilitate generalized enhancements make it a versatile tool in the realm of image editing, demonstrating the potential of language-driven image manipulation.

\paragraph{Frido}
Frido~\cite{Frido} adopts a detailed and nuanced approach to image processing, breaking down the input image into scale-independent quantized features. 
It utilizes a multi-scale MS-VQGAN to encode the input image into a latent space. 
Frido then conducts diffusion within this latent space through a sophisticated coarse-to-fine gating mechanism. 
This method allows for intricate control over the image generation process, resulting in high-quality and detailed outputs.

\paragraph{Versatile Diffusion and UniDiffuser}
Traditionally, diffusion models like Versatile Diffusion~\cite{Versatile-Diffusion} have focused on single-task workflows, often limited to generating one type of output based on a specific context. 
UniDiffuser~\cite{UniDiffuser} broadens this scope by introducing a unified diffusion model framework. 
It employs the Transformer as the denoising network backbone to handle multimodal data distributions, covering a range of tasks including text-to-image, image-to-text, and joint image-text generation. 
This innovative approach uses pre-trained encoders to map images and texts into a latent space, weaving together their embeddings to guide the generation of diverse modalities within a transformer-based diffusion process.

\paragraph{KNN-Diffusion}
Developing text-to-image models often face challenges due to the need for large datasets of text-image pairs, especially in domains with limited data availability. 
KNN-diffusion~\cite{knndiffusion} addresses this challenge by utilizing large-scale retrieval methods, primarily efficient k-Nearest-Neighbors (kNN) algorithms. 
This technique enables the training of compact and efficient text-to-image diffusion models without relying on text inputs. 
It can generate out-of-distribution images by altering the retrieval database during inference and perform text-driven local semantic edits while preserving object identities. 
This approach opens new possibilities for text-to-image synthesis, particularly in data-constrained environments.

\paragraph{DALL-E 3}
A persistent challenge in previous diffusion models is the controllability of image generation systems, which often overlook the words, word ordering, or meaning in a given caption. This challenge is commonly referred to as ``prompt following"\footnote{https://cdn.openai.com/papers/dall-e-3.pdf}. DALL-E 3\footnote{https://cdn.openai.com/papers/dall-e-3.pdf} presents a novel perspective on this issue by highlighting the potential improvement in prompt-following abilities of text-to-image models through training on highly descriptive generated image captions. The fundamental premise of their approach is grounded in the hypothesis that the deficiencies in prompt following arise from the presence of noisy and inaccurate image captions within the training dataset. To address this, DALLE3 takes a proactive step by training a robust image captioner and using it to recaption the
training dataset. Subsequently, several text-to-image models are trained using this refined dataset. The key finding is the consistent enhancement in prompt-following abilities observed in models trained on these synthetic captions.  The approach thus introduces a novel text-to-image generation system that stands out for its improved prompt-following characteristics.

\paragraph{Comprehensive Diffusion Model Overview}
The field of diffusion models has expanded considerably, with several noteworthy models contributing to its growth. 
Models like Unitune~\cite{unitune}, DiffusionCLIP~\cite{DiffusionCLIP}, Imagic~\cite{Imagic}, DreamBooth~\cite{DreamBooth}, eDiff-I~\cite{eDiff-I}, and ERNIE-ViLG 2.0~\cite{ernievilg} have each made significant strides in enhancing the capabilities and applications of diffusion models. 
These models explore various aspects of image synthesis, manipulation, and enhancement, thereby enriching the diversity and utility of diffusion-based approaches in visual content generation.

\paragraph{The Pinnacle of Diffusion Models as Foundation Models}
The widespread integration of diffusion models into the foundation model framework underscores their versatility and effectiveness. 
As likelihood-based models, diffusion models excel by avoiding the common pitfalls of mode-collapse and training instabilities often associated with GANs. 
Unlike GANs, which can struggle with maintaining diversity in generated content, diffusion models adeptly capture the complex distributions found in natural images. 
This is achieved without necessitating the use of an excessively large number of parameters, a contrast to the approach commonly seen in Autoregressive models~\cite{VQ-VAE-2}. 
Furthermore, the ability of diffusion models to scale effectively while preserving a stable training objective positions them as a vital and robust component in the realm of VFMs. 
Their consistent performance, coupled with the ability to handle complex image distributions, marks them as a cornerstone technology in the current AI landscape, particularly in tasks requiring high-fidelity image generation and manipulation.

\subsection{Benchmark Datasets in Visual Foundation Model Research}
Benchmark datasets are indispensable in AI research, providing essential platforms for evaluating and benchmarking the performance of models. 
Within the realm of visual foundation models, these datasets are crucial for both quantitative and qualitative assessments, addressing factors such as image fidelity and the congruence of text-image synthesis. 
The selection of an appropriate dataset is largely dependent on the specific task and the intricacies involved in text-to-image synthesis. A comprehensive list of datasets discussed in this section can be found in Table~\ref{tab:datasets}.

\paragraph{Prominent Datasets for Text-to-Image Synthesis}
Three datasets commonly used in text-to-image synthesis are the Oxford-120 Flowers dataset~\cite{Oxford-120-Flowers}, the CUB-200 Birds dataset~\cite{Caltech-UCSD-Birds}, and the MS COCO dataset~\cite{MS-COCO}. 
The MS COCO dataset is particularly noteworthy for its extensive content, supporting a wide range of tasks such as object detection, segmentation, key-point detection, and captioning. 
This dataset features an impressive collection of over 328K images, making it a valuable resource for comprehensive evaluation.

\paragraph{Datasets Designed for Human Evaluation}
Certain datasets are specifically designed for human evaluation, where human raters are employed to qualitatively assess and compare the output of various models. 
DrawBench~\cite{imagen}, PartiPropts~\cite{parti}, and UniBench~\cite{Upainting} are examples of such datasets. 
UniBench, for instance, offers an evaluation framework encompassing a range of scenes and includes prompts in both Chinese and English, catering to different levels of complexity. 
PartiPropts presents a distinct challenge with a diverse array of over 1600 English prompts, each with an associated ``challenge dimension'' to highlight the complexity involved in the prompt.

\paragraph{Evaluating Visual Reasoning and Biases}
The PaintSkills dataset~\cite{Dall-eval} focuses on evaluating visual reasoning capabilities and potential social biases in models, in addition to assessing image quality and text-image alignment. This dataset represents a critical step in understanding and improving the societal impacts and cognitive abilities of visual foundation models.

\paragraph{Visual Genome Dataset for Visual Question Answering}
The Visual Genome dataset is a significant resource for visual question-answering tasks. It consists of 101,174 images sourced from MS COCO and includes over 1.7 million QA pairs, averaging 17 questions per image. The dataset covers a wide spectrum of question types and is augmented with detailed annotations of objects, attributes, and relationships. The Parti model, for instance, demonstrated impressive performance on this dataset, achieving a Fréchet Inception Distance (FID) score of 3.22.

In conclusion, while automated metrics are widely used for model evaluations~\cite{parti, imagen, EntityDrawBench, Dall-eval, petsiuk2022human, liao2022artbench}, the development and application of specialized datasets are crucial for comprehensive model assessments. These datasets enable a deeper understanding of model capabilities and limitations, significantly contributing to the advancement of AI and visual foundation models.

\begin{table*}[!h]
    \centering
    \caption{Representative text-to-image datasets}
    \label{tab:datasets}
    \resizebox{\textwidth}{!}{%
        \begin{tabular}{
        p{0.1\linewidth} 
        p{0.27\linewidth}
        p{0.1\linewidth}
        p{0.21\linewidth} 
        p{0.21\linewidth} 
        p{0.1\linewidth}
        p{0.21\linewidth}
        }
            \toprule
            \textbf{Task} & 
            \textbf{Dataset} & 
            \textbf{Year} & 
            \textbf{Training} & 
            \textbf{Validation} & 
            \textbf{Testing} & 
            \textbf{Usage Examples} \\
            \midrule
            Text to Image Synthesis & MSCOCO~\cite{MS-COCO} & 2014 & 82,783 & 40,504 & 40,775 & Model Evaluation \\
            & CUB-200 Birds~\cite{Caltech-UCSD-Birds} & 2010 & 8,855 & - & 2,933 & Model Evaluation \\
            & Oxford-120 Flowers~\cite{Oxford-120-Flowers} & 2008 & 1,030 & 1,030 & 6,129 & Model Evaluation \\
            & LAION-400M~\cite{LAION-400M} & 2021 & 400M & - - & - - & Large Scale Training Datasets \\
            & CC3M~\cite{CC3M} & 2018 & 3,318,333 & 28,355 & 22,530 & Large Scale Training Datasets \\
            & CC12M~\cite{CC12M} & 2021 & 12,423,374 & - - & - - & Large Scale Training Datasets \\
            \midrule
            prompts & DrawBench~\cite{imagen} & 06/2022 & User Preference Rates & Fidelity, Alignment & 200 & Model Evaluation \\
            & PartiPrompts~\cite{parti} & 06/2022 & Qualitative, User Preference Rates & Fidelity, Alignment & 1,600 & Model Evaluation \\
            & UniBench~\cite{Upainting} & 10/2022 & User Preference Rates & Fidelity, Alignment & 200 & Model Evaluation \\
            & PaintSKills~\cite{Dall-eval} & 02/2022 & Statistics & Visual Reasoning Skills, Social Biases & 145 & Model Evaluation \\
            & EntityDrawBench~\cite{EntityDrawBench} & 09/2022 & Human Rating & Entity-centric Faithfulness & 250 & Model Evaluation \\
            & Multi-Task Benchmark~\cite{petsiuk2022human} & 11/2022 & Human Rating & Various Capabilities & N/A & Model Evaluation \\
            \bottomrule
        \end{tabular}%
    }
\end{table*}

\subsection{Evaluation Metrics for GVFMs}

Evaluating visual foundation models demands a careful balance between automated quantitative metrics and human judgment. This dual approach provides a comprehensive assessment, encompassing various aspects of model performance such as image quality, diversity, and alignment with textual descriptions.

\paragraph{Automated Metrics for Image Quality}
\begin{itemize}
    \item \textbf{Fréchet Inception Distance (FID)}: FID quantifies the similarity between distributions of real and generated images by analyzing features extracted using a pre-trained Inception model~\cite{heusel2018gans}. A lower FID score indicates that the generated images closely resemble real images in terms of both visual quality and diversity.

    \item \textbf{Inception Score (IS)}: The Inception Score evaluates the diversity and quality of generated images based on the classification confidence of a pre-trained Inception model~\cite{salimans2016improved}. A higher IS denotes greater diversity and image quality. However, it's crucial to acknowledge that IS sometimes presents a trade-off with FID. Despite its straightforward application, IS may not effectively detect model overfitting or intra-domain variations and can be sensitive to noise.

    \item \textbf{Precision}: This metric assesses the correctness of generated images in specific contexts, such as image classification, using a pre-trained classifier~\cite{sajjadi2018assessing,kynkäänniemi2019improved}. A higher precision score suggests that the generated images closely resemble authentic images in relation to their intended use.
\end{itemize}

It is important to note that while FID is generally more comprehensive than IS in assessing the quality range of generated images, it assumes a Gaussian distribution of image features, which may not always be accurate.
Additionally, diversity-focused metrics like LPIPS can erroneously assign high scores to unrealistic images due to their sole focus on image diversity.

\paragraph{Metrics for Image-Text Alignment}
\begin{itemize}
    \item \textbf{CLIP Score:}
The CLIP Score assesses the semantic alignment between images and captions~\cite{CLIPScore}. It quantifies the degree of semantic congruence between an image and its corresponding text. A higher CLIP score indicates better alignment, often showing a strong correlation with human evaluations.

    \item \textbf{Human Evaluation:}
Human evaluations are integral for validating the effectiveness and relevance of automated metrics. Studies~\cite{Upainting,imagen,parti,EntityDrawBench,petsiuk2022human} have incorporated user studies to compare the performance of generated images against human perceptions. For instance, in~\cite{imagen}, participants rated generated images on alignment and quality. They compared model-generated images with reference images to determine photorealism, using the preference rate as a benchmark. In alignment evaluations, human raters assigned scores to image-text pairs. Similarly,~\cite{petsiuk2022human} employed human assessment across varying levels of difficulty to compare models like Stable Diffusion and DALL-E 2. This approach is especially valuable for evaluating subjective elements, such as the creative combination of objects or the avoidance of biases related to race or gender. In scenarios requiring common-sense understanding or sensitivity to societal biases, human evaluations are indispensable, providing insights that automated metrics may not capture.
\end{itemize}

\subsection{Commercialized Products with GVFMs}

Applications of GVFMs in image generation have seen remarkable advancements, leading to the development of various products that excel in creating visually stunning content from textual prompts. As shown in Table~\ref{table: GVFM products}, these products are accessible across various platforms, including web and mobile, significantly contributing to the dynamic landscape of AI-driven visual content creation.

\paragraph{DALL-E and DALL-E 2}
DALL-E~\cite{DALLE}, a trailblazer in the text-to-image generation domain, initially utilized GANs to produce images with a distinct, somewhat cartoonish quality against simple backdrops. 
The advent of DALL-E 2~\cite{DALLE2product} marked a significant shift to a diffusion model framework, greatly enhancing the system's capability to generate photorealistic images with complex details. 
This evolution from DALL-E to DALL-E 2 demonstrates a profound improvement in visualizing diverse concepts with greater realism and precision.

\paragraph{Midjourney}
Developed by Midjourney Labs, Midjourney AI\footnote{https://www.midjourney.com/home/} specializes in generating surreal imagery based on textual inputs. 
It offers various model versions, each designed to cater to different artistic styles~\cite{midjourney-model-versions}.
For instance, Model Version 5.1 focuses on user-friendly interactions, producing clear and coherent images from succinct prompts and incorporating unique features like pattern repetition. 
In contrast, Model Version 5 tends to produce more photorealistic outputs, requiring more detailed prompts to attain specific visual outcomes.

\paragraph{Stable Diffusion}
Stable Diffusion~\cite{stable-diffusion-announcement} takes an open-source approach, providing users with a wide array of models and datasets for immediate art creation. 
Leveraging the advancements in vision models such as LDM~\cite{stable_diffusion}, DALL-E 2~\cite{DALLE2}, and Imagen~\cite{imagen}, Stable Diffusion stands out in converting text into detailed and vibrant imagery. 
Stability AI is currently beta testing an advanced version, Stable Diffusion XL~\cite{stable-diffusion-XL}, on platforms like DreamStudio. 
This new iteration aims to enhance user experience with additional functionalities, including inpainting, outpainting, and image-to-image transformations.

\paragraph{Adobe Firefly}
Adobe Firefly\footnote{Adobe Firefly https://firefly.adobe.com/}, distinct from other products, sets itself apart by offering a wide range of image customization features. 
It goes beyond basic text-to-image synthesis, allowing users to add or remove objects from images based on text descriptions, apply various text effects, and experiment with diverse color palettes for vector art. 
Adobe Firefly stands as a versatile and powerful tool, particularly for artists and designers, by integrating these comprehensive creative functionalities.

To assist readers interested in exploring this evolving field, Table~\ref{table: GVFM products} provides a detailed overview of key online platforms proficient in generating images from text prompts. This table serves as a valuable resource for navigating the extensive array of tools available in this rapidly developing domain of AI image generation.

\begin{table*}[!h]
\centering
\caption{Generative Visual Foundation Models Products }
\label{table: GVFM products}
\resizebox{\linewidth}{!}
{
\begin{tabular}{
p{0.2\linewidth} 
p{0.26\linewidth} 
p{0.28\linewidth} 
l 
l  
p{0.2\linewidth}}
        \toprule
        \textbf{Products} & \textbf{Base Models} &\textbf{Features} & \textbf{Links} & \textbf{Platforms} & \textbf{Free}\\ 
        \midrule
        DALL-E 2 & DALL-E 2~\cite{DALLE2} &text to image &\href{https://openai.com/dall-e-2}{DALL-E 2} & Web & No\\
        Openjourney & stable diffusion~\cite{stable_diffusion},  \href{https://huggingface.co/prompthero/openjourney}{openjourney} (an open source Stable Diffusion fine tuned model on Midjourney images)&text to image&\href{https://openjourney.art/}{Openjourney}&Web&Yes\\
        Midjourney & - - & text to image& \href{https://www.midjourney.com/home/}{Midjourney} & Web (Discord)&No\\
        Firefly & - - & text-to-image generation, image expansion, vector recoloring, text effects, inpainting, sketch-to-image(some in development)& \href{https://firefly.adobe.com/inspire/images}{Firefly}&Web & Yes\\
        Stable Diffusion &Stable Diffusion~\cite{stable_diffusion}& text to image &\href{https://stablediffusionweb.com\#demo}{stable diffusion}& Web& Yes\\
        Playground AI &stable diffusion~\cite{stable_diffusion}, DALL-E~\cite{DALLE}&text to image&\href{https://playgroundai.com/login?redirect=/create?}{Playground AI}&Web& Free up to 1000 images/ day\\
        Mage space & - - & text to image &\href{https://www.mage.space/}{Mage space}&Web&No\\
        Leonardo AI  & - - & primarily developed to generate game assets&\href{https://leonardo.ai/}{Leonardo AI}& Web& Yes\\
        Shutterstock AI & - - &text to image, various visual styles &  \href{https://www.shutterstock.com/}{Shutterstock AI} & Web, Android, iOS\\
        fotor AI & stable diffusion~\cite{stable_diffusion} & text to image &\href{https://www.fotor.com/features/ai-image-generator/}{fotor AI} &Web, Android, iOS &free for basic edition \\
        StarryAI & Stable Diffusion~\cite{stable_diffusion} & text to image &\href{https://starryai.com/}{StarryAI} & Android, iOS &Generate up to 25 images for free daily and without watermarks\\ 
        Craiyon & DALL-E Mega, DALL-E Mini & text to image & \href{https://www.craiyon.com/}{Craiyon} &Web &have free editions \\
        NightCafe & Stable Diffusion~\cite{stable_diffusion}, DALL-E 2~\cite{DALLE2},CLIP-Guided Diffusion, VQGAN+CLIP, and Style Transfer &  Style Transfer(create masterpieces modeled on old artworks); CLIP-Guided Diffusion (artistic images); VQGAN+CLIP (generate beautiful sceneries) &\href{https://creator.nightcafe.studio/studio}{NightCafe} & Web & Unlimited base Stable Diffusion generations, plus daily free credits to use on more powerful generator settings.\\
        DeepAI & stable diffusion~\cite{stable_diffusion} & style& \href{https://deepai.org/}{DeepAI} & Web & free for basic version\\
        Wombo & - - &style&\href{https://dream.ai/create}{Wombo}&Android, iOS & free for basic edition\\
        Baidu Yige & ERNIE-ViLG 2.0~\cite{ernievilg} &text to images, chinese prompts and chinese style images, image editing, and one-click video production &\href{https://yige.baidu.com\#/}{Baidu Yige}& Web& free\\
        Freehand & - - & chinese prompts, text to image&\href{https://freehands.cn/}{Freehand}&Web& free\\
        Playground AI & stable diffusion~\cite{stable_diffusion} &text to image &\href{https://playgroundai.com/login}{Playground AI} &Web &free for 1,000 images per day\\
        Bing Image Creator & DALL-E~\cite{DALLE} & Integrated with Bing Chat & \href{https://www.bing.com/create}{Bing Image Creator} & Web, Android, iOS & Yes\\
        Lexica & fine tuned Stable Diffusion~\cite{stable_diffusion} & text to image &\href{https://lexica.art/}{Lexica}& Web & No\\
        Dreamstudio &stable diffusion~\cite{stable_diffusion}&text to images&\href{https://beta.dreamstudio.ai/generate}{Dreamstudio}& Web & No \\
        Canva & - - & text to image, Various art styles available (Watercolor, Filmic, Neon, Color Pencil, Retrowave etc) &\href{https://www.canva.com/ai-image-generator/}{Canva} & Web & Text to Image is available to free users who can access up to 50 lifetime queries\\
        DRAI &Kandinsky, Openjourney, Stable Diffusion 1.5, Stable Diffusion 2.0, Anything 3 and Anything 4&text to image, inpainting&\href{https://beta.dreamstudio.ai/generate}{DRAI}&iOS & No\\
        RunwayML & Gen-1~\cite{Gen-1}, Gen-2&text to video, video to video, image to video, text to image  &\href{https://runwayml.com/}{RunwayML}&Web, iOS&free for basic edition \\
            \bottomrule
        \end{tabular}
}
\end{table*}

\begin{table*}[!h]
\centering
	\caption{Summary of generative visual foundation model.} \label{tab:generative_models}  
    \resizebox{\linewidth}{!}
    {
    \begin{tabular}{
    p{0.15\textwidth} 
    p{0.14\textwidth} 
    p{0.15\textwidth} 
    p{0.1\textwidth} 
    p{0.2\textwidth} 
    p{0.2\textwidth} 
    p{0.3\textwidth} 
    p{0.2\textwidth} 
    p{0.1\textwidth}
    }

        \toprule
       \textbf{Name} &\textbf{Year(of first known publication)} & \textbf{Application} &\textbf{Base Model} &\textbf{Evaluation Metrics} &\textbf{Resource}& \textbf{Num. Params} & \textbf{Training Time} & \textbf{Github Link}\\
       
      \hline


       AttnGAN~\cite{xu2017attngan} &11/2017 & Text to image & - - &IS, R-precision & - - & - - & - - & - -\\ 

        BigGan~\cite{BigGAN} &09/2018 &Text to image & - - &FID, IS & a Google TPU v3 Pod, with the number of cores proportional to the resolution: 128 for $128^2$, 256 for $256^2$, and 512 for $512^2$ & BigGAN-deep G and D : 50.4M and 34.6M parameters respectively, original BigGAN models : 70.4M and 88.0M parameters. ($128^2$ resolution) & Training takes between 24 and 48 hours for most models &\href{https://github.com/ajbrock/BigGAN-PyTorch}{BigGan}\\
        
       StyleGAN~\cite{styleGAN} & 12/2018 & Text to image & - - & FID, Path length & an NVIDIA DGX-1 with 8 Tesla V100 GPUs & - - & one week on an NVIDIA DGX-1 with 8 Tesla V100 GPUs & \href{https://github.com/NVlabs/stylegan}{StyleGAN}\\

       StyleGAN2~\cite{styleGAN2} & 12/2019 & Text to image & StyleGAN & FID, Path length, Precision, Recall &NVIDIA DGX-1& generator(25M $\rightarrow$  30M), discriminator(24M $\rightarrow$ 29M) for resolutions $64^2$–$1024^2$ &trains at 37 images per second on NVIDIA DGX-1 with 8 Tesla V100 GPUs ($1024 \times 1024$ resolution) & \href{https://github.com/NVlabs/stylegan2}{StyleGAN2} \\

       VQ-GAN~\cite{VQ-GAN} & 12/2020 & Text to image & - - & FID and IS & trained with a batch-size of at least 2 on a GPU with 12GB VRAM, generally train on 2-4 GPUs with an accumulated VRAM of 48 GB & the number of transformer parameters varies from 85M to 307M for different experiments & -- & \href{https://github.com/CompVis/taming-transformers}{VQ-GAN} \\

       DALL-E~\cite{DALLE} & 02/2021 & Text to image & - - &IS, FID, human evaluation& 16 GB NVIDIA V100 GPUs & - - & - - &\href{https://github.com/openai/DALL-E}{DALL-E}\\

        ADM~\cite{DiffusionBeatGANs} &05/2021&Text to image& - -& - -& - - & LSUN(552M), ImageNet 64(296M), ImageNet 128(422M), ImageNet 256(554M), ImageNet 512(559M)& - -& - -\\

        VQ-diffusion~\cite{VQ-diffusion} & 11/2021 & Text to image & - - & - - & - - & 1) VQ-Diffusion-S (Small)
        : 34M parameters. 2) VQ-Diffusion-B (Base) : 370M parameters & - -& \href{https://github.com/cientgu/VQ-Diffusion}{VQ-diffusion}\\
       
        GLIDE~\cite{Glide}&12/2021&Text to image& - - & - - & - - & 3.5B ($64 \times 64$ resolution(2.3B for visual encoding, 1.2B for textual)) + 1.5B ($256 \times 256$) & - -&\href{https://github.com/openai/glide-text2im}{GLIDE}\\

        LDM~\cite{stable_diffusion}&12/2021&Text to image& - -& - - &a single NVIDIA A100 & unconditional LDMs (LDM-1: 270M, LDM-2: 265M, LDM-4: 274M, LDM-8: 258M, LDM-16: 260M, LDM-32:258M), conditional LDMs(Text-to-Image: 1.45B, Layout-to-Image trained on OpenImages: 306M, Layout-to-Image trained on COCO: 345M, Class-Label-to-Image: 395M, Super Resolution: 169M, Inpainting: 215M, Semantic-Map-to-Image: 215M) & - -&\href{https://github.com/CompVis/latent-diffusion}{LDM}\\
        
        DALL-E 2~\cite{DALLE2}&04/2022&Text to image& - - & - - & - - & 3.5B ($64 \times 64$ resolution) + 700M ( $256 \times 256$) + 300M ($1024 \times 1024$ resolution) & - - & - -\\
        
        Imagen~\cite{imagen}&06/2022&Text to image& - - & FID, human evaluation & 256 TPU-v4 chips for $64 \times 64$ model, and 128 TPU-v4 chips for both super-resolution & 2B ($64 \times 64$ resolution) + 600M ($256 \times 256$ resolution) + 300M ($1024 \times 1024$ resolution) & - - & - -\\

        Parti~\cite{parti}&06/2022 &Text to image & - -& &CloudTPUv4 hardware&20B &- -&\href{https://github.com/google-research/parti}{Parti}\\
        \bottomrule
        \end{tabular}%
    }
\end{table*}

\section{Discriminative Visual Foundation Models}

\subsection{Definition and Formulation}

\paragraph{Visual Discriminative Model}
In the landscape of VFMs, discriminative models stand in contrast to their generative counterparts. 
While generative models focus on capturing the underlying data distribution to create new samples, discriminative models specialize in learning the decision boundaries between various classes or categories within a dataset. 
These models excel in predicting the correct label or class for a given input, rather than generating new data. 
Their primary role in computer vision is classification, where they are tasked with accurately categorizing input data into specific classes based on learned patterns and features. 
This functionality makes them essential for tasks such as image classification~\cite{ImageNet}, object detection~\cite{plummer2016flickr30k}, and image segmentation~\cite{long2015fully,kirillov2019panoptic}.

\paragraph{Discriminative Visual Foundation Models}
\textit{Discriminative Visual Foundation Models} (DVFM) have traditionally been designed to excel in specific tasks within the domain of computer vision. 
Pioneering models in this field, such as AlexNet~\cite{AlexNet}, ResNet~\cite{ResNet}, and Vision Transformer (ViT)~\cite{ViT}, have significantly advanced the capabilities of discriminative tasks in visual perception.
The advent of pre-trained Visual Language Models (VLMs) has marked a significant shift in the evolution of DVFMs. 
Models like CLIP~\cite{CLIP}, ALIGN~\cite{ALIGN}, Florence~\cite{Florence}, VLBERT~\cite{vlbert}, and X-LXMERT~\cite{xlxmert} exemplify this evolution. 
These models are designed to capture the complex interaction between visual and linguistic elements, elevating the potential of foundation models in discriminative tasks. 
Once trained, they utilize text prompts tailored for specific tasks to achieve zero-shot generalization, adapting to novel visual concepts and data distributions with ease. 
This versatility allows them to be applied in a wide range of applications, including but not limited to classification, retrieval, object detection, video comprehension, visual question answering, image captioning, and even facilitating certain aspects of image generation. 
These models, under the category of ``textually prompted models''~\cite{awais2023foundational}, represent a significant stride in integrating language understanding with visual perception, broadening the scope and effectiveness of discriminative visual foundation models.

\begin{figure}[h!]
 \centering
\includegraphics[width=0.6\linewidth]
{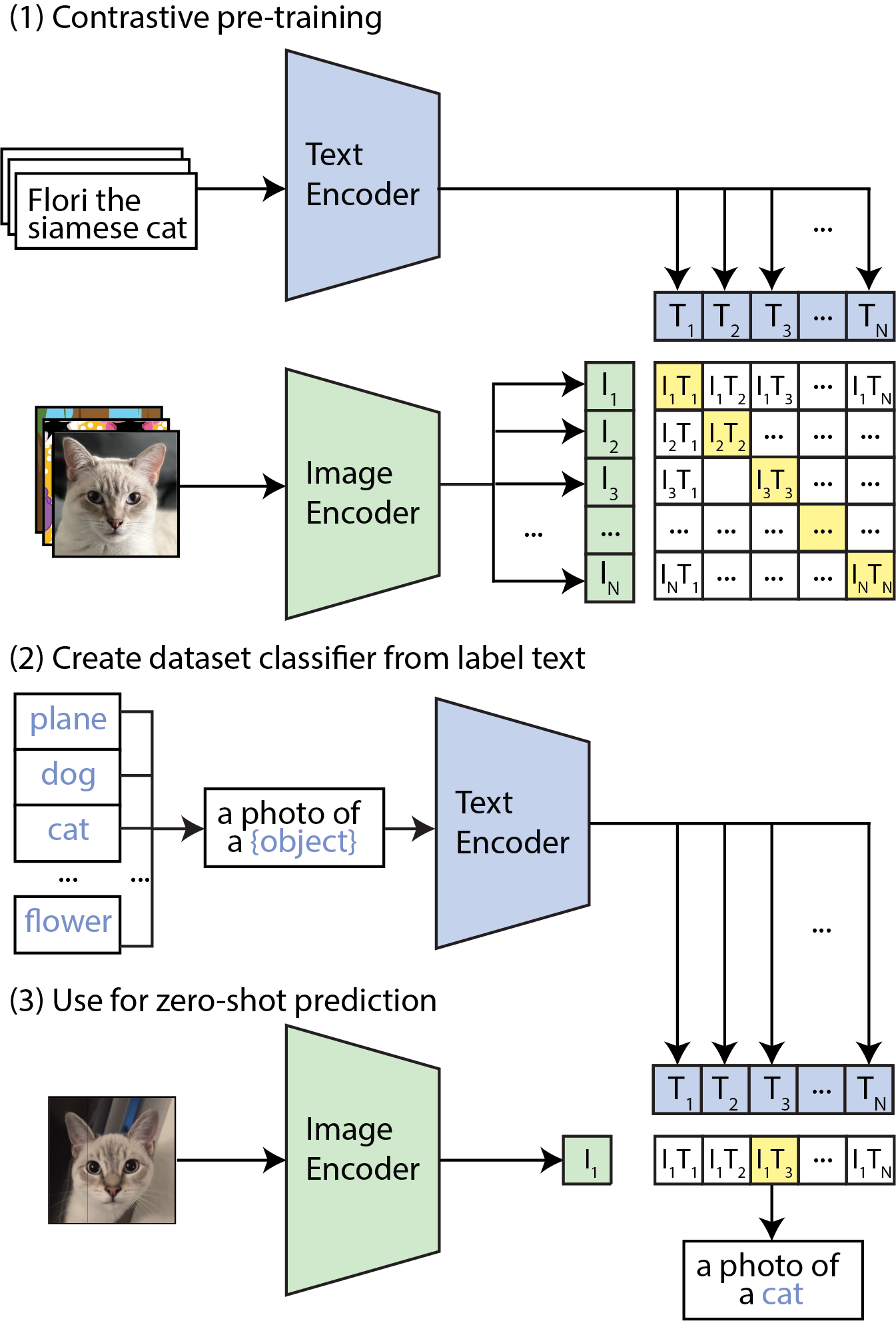}
\caption{This figure showcases the architecture of CLIP. 
The primary objective of CLIP is to accurately predict the correct pairings of a batch of (image, text) training examples. 
It achieves this by maximizing the alignment between the embeddings of matching image and text pairs while minimizing it for non-matching pairs. 
}
\label{fig:CLIP}
\end{figure}

\paragraph{Advancements in DVFMs}
The progress in DVFM has significantly revitalized the field of CV, yet these models often encounter limitations in interacting with humans, especially in scenarios demanding various human inputs beyond just language. To enable a smooth interaction between humans and AI, models need to be adept at not only understanding language prompts but also other types of prompts that can fill in missing information or resolve language-based ambiguities.
To address this, a new breed of models, referred to as promptable models~\cite{li2023multimodal}, has emerged. These models undergo pre-training on extensive datasets, engaging in tasks specifically designed to join or enhance language prompts with other types of prompts.  
These models not only respond to textual prompts but can also interpret visual cues such as points, bounding boxes, and masks.
Prominent examples include models like SAM~\cite{SAM} and SEEM~\cite{SEEM}, which highlight the continuous efforts to improve the generalization abilities of contemporary vision models.
A comprehensive summary of these discriminative visual foundation models is detailed in Table~\ref{tab:discriminative_models}.


\subsection{Techniques in DVFMs}

\subsubsection{Architectures for Learning Image Features}

\begin{figure}[h!]
 \centering
\includegraphics[width=0.5\linewidth]
{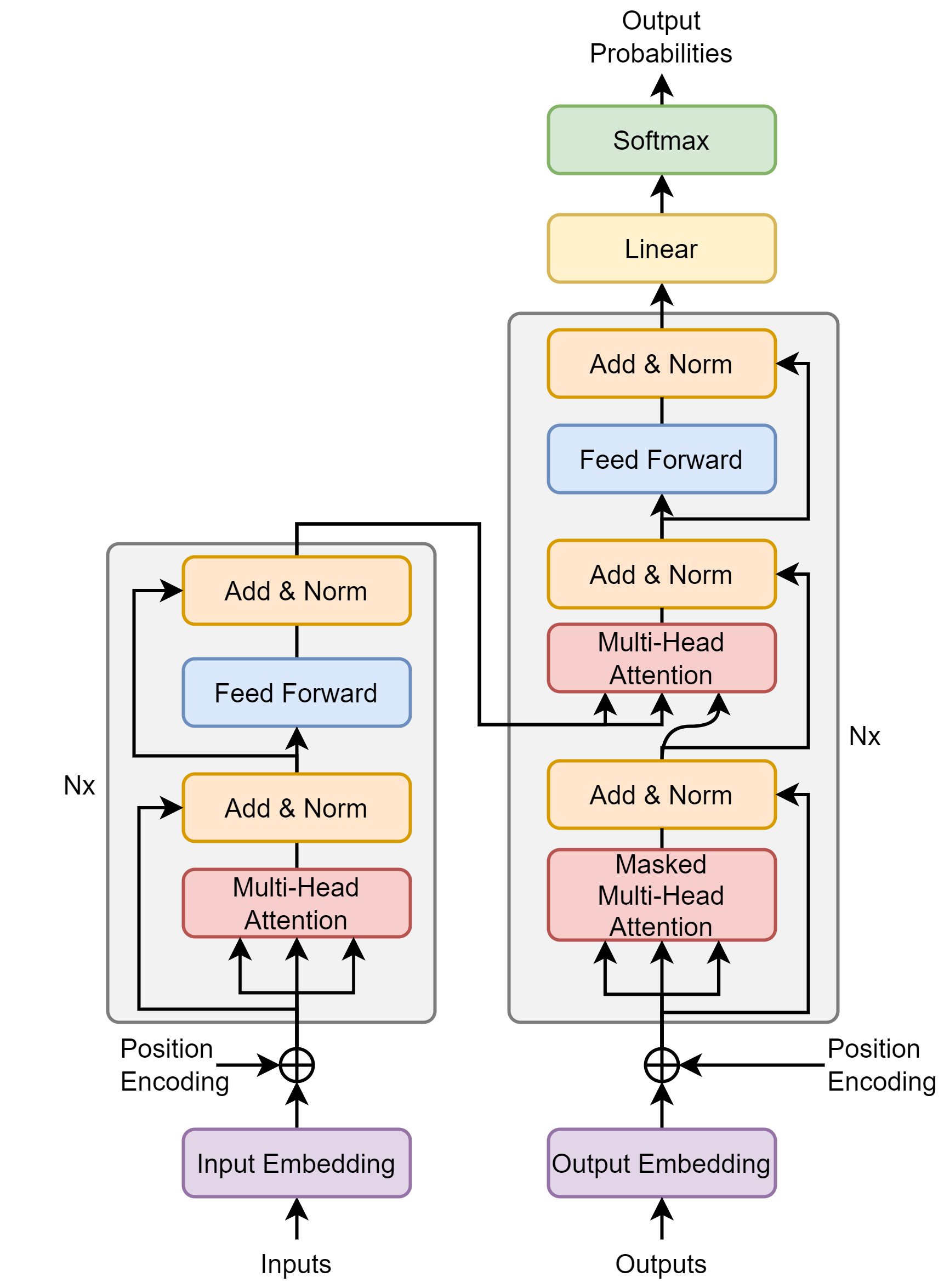}
\caption{Illustration of the Transformer, characterized by its innovative use of stacked self-attention and point-wise, fully connected layers. This diagram depicts both the encoder and decoder components of the Transformer. The encoder processes the input data through a series of self-attention and feed-forward layers, effectively encoding the input into a higher-level representation. The decoder, mirroring this structure, utilizes the encoded data to generate the final output. This architecture has been pivotal in advancing various fields, particularly in natural language processing and computer vision, due to its efficiency in handling sequential data and its ability to capture long-range dependencies within the input.}
\label{fig:transformer_framework}
\end{figure}

\paragraph{Transformers in Visual Recognition}
Transformers have gained prominence in visual recognition tasks, including image classification~\cite{ViT,liu2021swin}, object detection~\cite{carion2020endtoend,Deformable_DETR}, and semantic segmentation~\cite{zheng2021rethinking,xie2021segformer}. 
The Vision Transformer (ViT)~\cite{ViT} exemplifies the application of the standard Transformer architecture in image feature learning. 
In ViT, an image is divided into fixed-size patches, which are linearly projected and fed into a stack of Transformer blocks, each comprising a multi-head self-attention layer and a feed-forward network. 
Positional embeddings are added to maintain spatial information. 
Figure~\ref{fig:transformer_framework} illustrates the ViT framework. 
In DVFM studies like CLIP~\cite{CLIP} and SLIP~\cite{SLIP}, this architecture is slightly modified with the addition of an extra normalization layer before the Transformer encoder, showcasing the adaptability of Transformer models in handling complex visual tasks.
The Swin-Transformer~\cite{liu2021swin} is another milestone for computer vision. 
As a hierarchical Transformer, it adopts shifted windows for representation learning, allowing ViT-like architectures to generalize to higher resolution images. 

\subsection{Training Paradigms for DVFMs}

\subsubsection{Self-Supervised Learning}
Transformers have shown promising results in various CV tasks. 
However, they often require more training data compared to traditional CNNs. 
To overcome this data dependency, recent advancements have embraced self-supervised learning paradigms. Pioneering works like MoCo~\cite{MoCo} and SimCLR~\cite{SimCLR} utilize unsupervised pre-training followed by fine-tuning and prediction, leveraging unlabeled data to learn transferable representations. 
Various self-supervised training objectives, or pretext tasks, have been developed, including image inpainting~\cite{context_encoders}, masked image modeling~\cite{MAE,BEiT,SimMIM}, and contrastive learning~\cite{CLIP}. 
These approaches enable the learning of discriminative features without the need for labeled data during pre-training, leading to enhanced performance compared to supervised pre-training methods.

\subsection{Image Classification with DVFMs}
Image Classification involves categorizing images into predefined classes. 
The application of specific task DVFMs in image classification often involves textually prompted models built on pre-trained models. 
These models accomplish zero-shot image classification by comparing the embeddings of images with text, using ``prompt engineering'' to create task-specific prompts like ``\textit{a photo of a [label]}''. 

\subsubsection{Contrastive Learning-Based DVFMs}
We summarize prominent approaches within DVFMs is based on contrastive learning.

\paragraph{CLIP}
CLIP (Contrastive Language-Image Pre-training)~\cite{CLIP} employs image-text contrastive learning, maximizing cosine similarity between correct image-text pairs and minimizing it for incorrect ones. 
This method has widespread applications in both discriminative and generative tasks, as illustrated in Figure~\ref{fig:CLIP}.

\paragraph{ALIGN}
ALIGN~\cite{ALIGN}, unlike CLIP, utilizes large-scale, raw alt-text data for pre-training, demonstrating that effective visual and vision-language representations can be learned from less curated datasets.

\paragraph{Florence}
Addressing the limitations of image-to-text mapping models like CLIP and ALIGN, Florence~\cite{Florence} introduces a novel approach, featuring a two-tower architecture with a language transformer and a hierarchical Vision Transformer. 
It incorporates task-specific adapters, enhancing its applicability across various domains, including extending features temporally (from static images to videos) and modally (from images to language).

\subsubsection{Hybrid Contrastive and Generative Methods}

\paragraph{BLIP and BLIP2} 
BLIP~\cite{BLIP}, as well as its successor BLIP-2~\cite{BLIP2}, stands for Bootstrapping Language-Image Pre-training, emphasizing their commitment to achieving unified vision-language understanding and generation. Leveraging the power of Large Language Models (LLMs) and Vision Transformers (ViTs), both BLIP and BLIP-2 have demonstrated remarkable proficiency in various vision-language tasks, including image captioning, visual question answering, and image-text retrieval. 
The optimization strategy of BLIP~\cite{BLIP} revolves around three key objectives: image-text contrastive learning, image-text matching, and language modeling. 
It uses a Multimodal Encoder-Decoder (MED) structure that functions as a unimodal encoder, image-grounded text encoder, or decoder, depending on the task. This multi-task approach makes BLIP adaptable to a wide range of vision-language tasks. Building upon the success of BLIP~\cite{BLIP}, BLIP2~\cite{BLIP2} jointly optimize three pre-training objectives that share
the same input format and model parameters.  Moreover, BLIP-2 introduces a novel component known as a Querying Transformer (Q-Former). This trainable module plays a crucial role in bridging the gap between a frozen image encoder and a frozen LLM, enhancing the model's overall capabilities.

In summary, the evolution of DVFMs, particularly in the realm of image classification, has seen significant advancements through the integration of self-supervised learning and textually prompted models. 
%

\subsection{Image Segmentation with DVFMs}
Image segmentation, a core task in CV, involves dividing a digital image into distinct segments, assigning each pixel to specific classes or objects. 
This task has traditionally encompassed three primary types: semantic segmentation, instance segmentation, and panoptic segmentation. 
Semantic segmentation~\cite{chen2016semantic,chen2017rethinking,chen2018encoderdecoder} focuses on classifying each pixel into predefined semantic classes. 
Instance segmentation~\cite{Hafiz_2020,liu2018path,bolya2019yolact} takes this further by differentiating individual instances within the same class. 
Panoptic segmentation, as introduced by~\cite{kirillov2019panoptic}, integrates both semantic and instance segmentation for a comprehensive scene understanding. 
Additionally, related tasks such as edge detection~\cite{5557884}, superpixel segmentation~\cite{1238308}, object proposal generation~\cite{object_proposal}, and foreground segmentation~\cite{784637} expand the scope of segmentation in computer vision. 
The overarching aim of DVFMs in this context is to develop versatile models capable of adapting to a wide array of segmentation tasks.

\begin{figure}[h!]
\centering
\includegraphics[scale=0.5]
{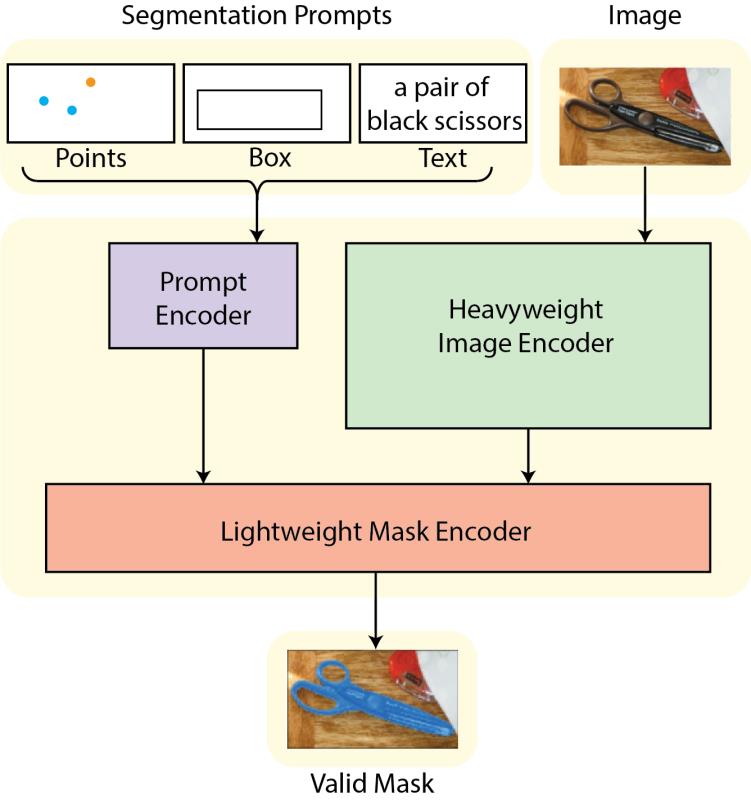}
\caption{Overview of the SAM. This model architecture is designed to effectively interpret and integrate information from two distinct sources: image embeddings and prompt embeddings. The image embeddings are generated by the image encoder, which processes the visual input, while the prompt encoder produces the prompt embeddings, derived from textual descriptions. These combined embeddings are then skillfully translated by the mask decoder into accurate segmentation masks. The figure illustrates this process using an example of scissors, showcasing SAM's capability to understand and segment complex objects.}
\label{fig:Segment_anything}
\end{figure}

\paragraph{Segment Anything Model (SAM)}
The Segment Anything Model (SAM)~\cite{SAM} exemplifies a prominent DVFM that achieves remarkable generalization capabilities. 
As a promptable model, SAM is pre-trained on a diverse dataset, employing tasks designed to enable zero-shot generalization. 
Its success is attributed to three core components: a task design facilitating zero-shot generalization, a model architecture that supports flexible prompting, and a data collection strategy tailored to empower the model and its intended tasks.
The SAM architecture, illustrated in Figure~\ref{fig:Segment_anything}, consists of three integral components: an image encoder, a prompt encoder, and a mask decoder. 
The image encoder processes the input image to produce image embeddings, while the prompt encoder generates prompt embeddings based on the input task description. 
The mask decoder then translates the combined information from these embeddings into valid segmentation masks. 
This model exemplifies the integration of textual prompts with visual cues, enabling it to adaptively segment images across various contexts and classes. 
The given figure, sourced from the original SAM paper, provides an overview of how SAM interprets an input image and corresponding prompt to produce accurate segmentation masks.


We delve deeper into the mechanisms and techniques employed within DVFMs like SAM, exploring how they redefine the landscape of image segmentation in computer vision.
\begin{enumerate}
    \item \textbf{Task Design}:
    Drawing inspiration from ``prompting'' techniques in NLP, a novel concept of promptable segmentation tasks is introduced. These tasks involve returning valid segmentation masks in response to diverse segmentation prompts, which may contain spatial or textual information identifying an object or feature within an image. This approach is not only utilized as a pre-training objective but also adapts to solve a variety of downstream segmentation tasks through innovative prompt engineering.
    
    \item \textbf{Model Architectures}:
    The architectural design, as depicted in Figure~\ref{fig:Segment_anything}, comprises three main components: an image encoder, a flexible prompt encoder, and an efficient mask decoder. 
    The image encoder utilizes a minimally adapted, MAE~\cite{MAE} pre-trained ViT~\cite{ViT} to handle high-resolution inputs. The prompt encoder represents sparse prompts (points, boxes, and text) using positional encodings combined with learned embeddings for each prompt type, and free-form text is processed using a CLIP-based text encoder~\cite{CLIP}. 
    Dense prompts (like masks) are embedded through convolutions and integrated with image embeddings. 
    The mask decoder is designed to map the combined embeddings and an output token to generate the mask. The model employs a loss function that is a linear combination of focal loss~\cite{lin2018focal} and dice loss~\cite{milletari2016vnet}, following the methodology used in end-to-end object detection models like~\cite{carion2020endtoend}.
    
    \item \textbf{Data Collection}:
    Addressing the challenge of limited public data, the researchers employed a training-annotation iterative process, creating a data engine for simultaneous model training and dataset construction. 
    This process encompasses three stages: assisted-manual, semi-automatic, and fully automatic. 
    Initially, SAM assists annotators in creating masks. 
    The semi-automatic stage involves SAM generating masks for some objects, with annotators focusing on the rest, thereby increasing mask diversity. 
    The final stage features SAM producing numerous high-quality masks per image using a grid of foreground points, thereby efficiently expanding the dataset.
\end{enumerate}

\paragraph{SEEM Model}
Building upon the diverse categories of prompts processed by SAM, SEEM~\cite{SEEM} takes a step further by encoding a wider array of prompts, including visual prompts (points, boxes, scribbles, masks), text prompts, and referring prompts (regions referred from another image). 
SEEM's ability to encode these varied prompts into a joint visual-semantic space empowers it with a robust zero-shot generalization capability, enabling it to effectively address unseen user prompts for segmentation tasks.

\paragraph{SegGPT}
SegGPT~\cite{SegGPT} adopts a novel perspective by treating segmentation as a universal format for visual perception, unifying various segmentation tasks under a generalist in-context learning framework. 
This model transforms different segmentation data into a uniform image format, with training formulated as an in-context coloring challenge. 
By employing a random coloring scheme, SegGPT encourages reliance on contextual information rather than specific colors to complete tasks. 
This versatile approach allows SegGPT to be applied across a wide range of segmentation tasks, such as few-shot semantic segmentation, video object segmentation, and panoptic segmentation, making it an exemplary DVFM.

\subsection{Object Detection with DVFMs}

Object detection, a pivotal task in computer vision, involves localizing and classifying objects in images or video sequences. 
This capability is crucial for numerous applications in the field. Recent developments have shown that pre-trained DVFMs, leveraging auxiliary datasets like Object365~\cite{Object365} and MDETR~\cite{kamath2021mdetr}, can achieve zero-shot prediction in object detection. 
These models compare embeddings of object proposals and corresponding texts, facilitating object identification without explicit training on those specific categories.

\paragraph{Open-Vocabulary Object Detection}
Traditional object detection models are limited to recognizing categories present in their training data. 
To address this limitation, zero-shot object detection methodologies~\cite{zero_shot_object,Rahman_Khan_Barnes_2020,zhu2020dont} have been proposed. 
These methods aim to generalize from seen categories (with bounding box annotations) to unseen categories. 
However, despite making significant progress, they lag behind fully-supervised approaches in terms of performance. 
Open vocabulary object detection~\cite{du2022learning} progresses this task further by incorporating image-text aligned training. In contrast to zero-shot object detection, this approach eliminates any constraints imposed by a predefined training vocabulary.
This approach utilizes the expansive vocabularies derived from textual data to enhance the model's ability to detect novel object categories. The emergence of large-scale image-text pre-training works~\cite{ALIGN,CLIP} has bolstered this approach. 
Recent methods~\cite{du2022learning,feng2022promptdet} have successfully integrated these pre-trained parameters into open-vocabulary detection, significantly improving performance and expanding detectable category vocabularies.

\begin{figure}[h!]
 \centering
\includegraphics[width=1.0\linewidth]
{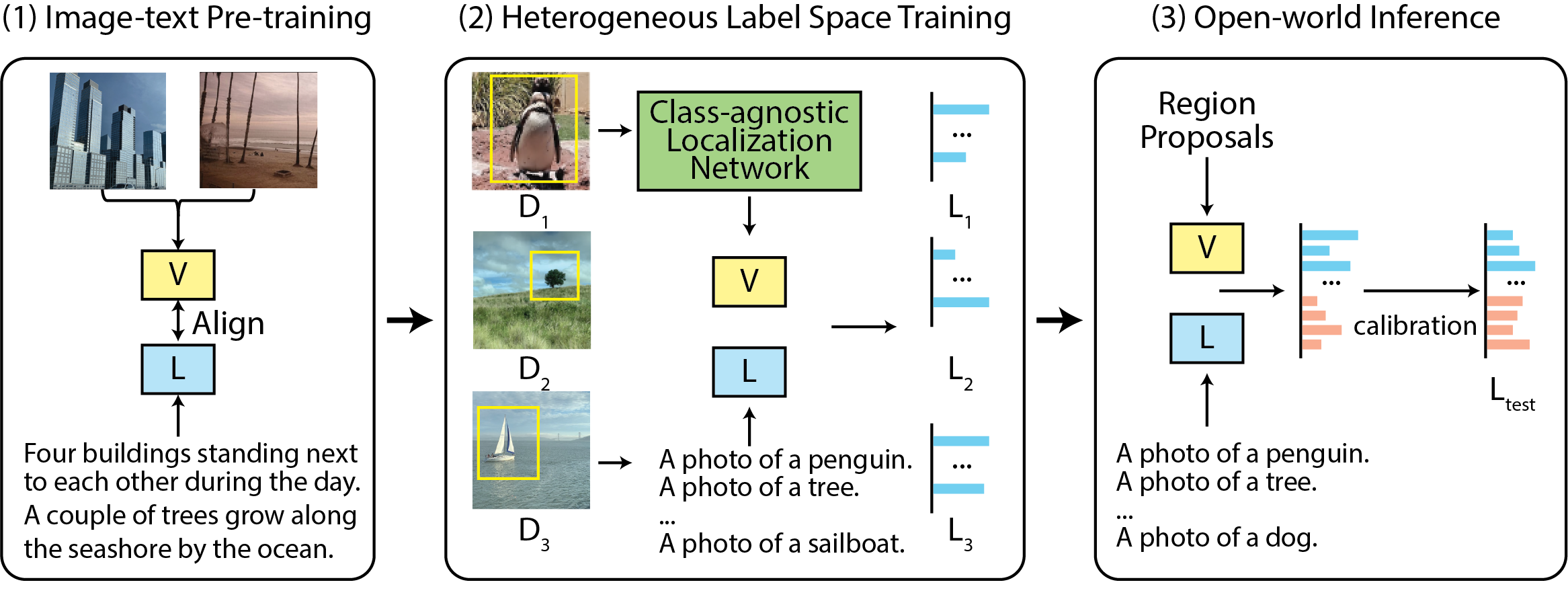}
\caption{An overview of the UniDetector~\cite{unidetector}, encompassing a three-stage process. 
Stage one involves image-text pre-training that seeks alignment between visual and textual representations. 
The second stage advances to training with a heterogeneous mix of labeled images, adopting a decoupled approach to enhance generalization across diverse label spaces. 
The final stage employs probability calibration techniques to balance the model's performance between seen and unseen classes, ensuring robust open-world inference capabilities. 
This figure illustrates how the UniDetector leverages large-scale pre-trained models to facilitate universal object detection, effectively scaling to a broad spectrum of object categories.
}
\label{fig:UniDetector}
\end{figure}

\paragraph{The UniDetector Framework}
The UniDetector~\cite{unidetector} framework specifically addresses the universal object detection challenge. As outlined in Figure~\ref{fig:UniDetector}, this framework encompasses three primary stages: large-scale image-text aligned pre-training, heterogeneous label space training, and open-world inference. 
Utilizing pre-training models like RegionCLIP~\cite{RegionClip}, UniDetector effectively generalizes to a wide range of objects, balancing detection of both seen and unseen classes. 
The heterogeneous label space training diverges from conventional object detection methodologies, which typically focus on a single dataset. 
Instead, UniDetector trains on diverse image sources with varying label spaces, essential for its universality. 
By decoupling the proposal generation and RoI classification stages, rather than training them jointly, it further enhances its ability to generalize to novel categories. 
Remarkably, with training on about 500 classes, UniDetector can detect over 7,000 categories. 
This impressive generalizability positions UniDetector as a leading candidate in the realm of DVFMs, paving the way for more adaptive and comprehensive object detection systems.

\subsection{Datasets}
\textbf{Datasets for Pre-training DVFMs}
\begin{itemize}
    \item \textbf{Microsoft COCO Captions}~\cite{chen2015microsoft}: This dataset is a collection of image-caption pairs with 567K captions for more than 113K images. Images are sourced from Flickr based on 80 object categories and varied scene types. Captions undergo tokenization via Stanford PTBTokenizer, and each image is associated with 5 (MS COCO c5) or 40 (MS COCO c40) captions. 

    \item \textbf{Visual Genome}~\cite{krishna2017visual}: Offering a dense annotation of images, Visual Genome includes region descriptions, objects, attributes, relationships, and related question-answer pairs for over 100K images, providing a rich resource for vision-language pretraining.

    \item \textbf{Conceptual Captions (CC3M and CC12M)}: CC3M~\cite{CC3M} and CC12M~\cite{CC12M} datasets significantly expand the image-text pair collection, with 3.3M and 12M pairs respectively. CC3M focuses on cleaned and hypernymized Alt-texts, while CC12M extends the dataset with relaxed filtering criteria.

    \item \textbf{CLIP Dataset}~\cite{CLIP}: Introduced by CLIP, this extensive dataset comprises 400M image-text pairs gathered from various internet sources. It uses 500k search queries to encapsulate a broad visual concept spectrum, ensuring a balanced representation across queries.

    \item \textbf{ALIGN Dataset}~\cite{ALIGN}: ALIGN presents 1.8B noisy image-text pairs, adopting a simplified frequency-based filtering approach from the extensive data collection methodology of Conceptual Captions, creating a dataset magnitudes larger in scale.

    \item \textbf{Wikipedia-based Image Text (WIT) Dataset}~\cite{srinivasan2021wit}: This Wikipedia-sourced dataset features 37.6M image-text pairs across 100+ languages. It provides a multilingual dataset with varied text types, such as references, attributions, and Alt-text descriptions, linked to corresponding images.

    \item \textbf{WenLan}~\cite{huo2021wenlan}: As a large Chinese image-text dataset, WenLan comprises 30 million image-text pairs across diverse topics. It employs topic models to ensure topic distribution balance and content quality.

    \item \textbf{RedCaps}~\cite{desai2021redcaps}: Sourced from Reddit, RedCaps curates a multimodal dataset with 12M image-text pairs. The collection process involves selective subreddit curation and stringent caption cleaning to ensure data relevance and quality.

    \item \textbf{LAION-5B}~\cite{schuhmann2022laion-5b}: This vast dataset encompasses 5.85B image-text pairs with multilingual support. To maintain data quality, a CLIP model filters out pairs below a certain similarity threshold, ensuring relevance between image and text pairings.
\end{itemize}


\subsubsection{Image Classification Datasets}
The realm of image classification encompasses a spectrum of tasks that vary in specificity and complexity. 
Foundational datasets such as ImageNet~\cite{ImageNet} serve general classification objectives, while specialized datasets like Oxford-IIIT PETS~\cite{cats_and_dogs} target fine-grained recognition tasks, focusing on detailed distinctions within a narrower domain, such as different breeds of cats and dogs. 
For more abstract categorization, such as texture classification, other curated datasets are utilized. The array of image classification datasets, each tailored to a particular scope of visual recognition, is systematically detailed in the first section of Table~\ref{tab:discriminative-datasets}.

\subsubsection{Image Segmentation Datasets}
The image segmentation discipline is subdivided based on the granularity of the segmentation task—whether it is semantic, where the aim is to label each pixel with a class; instance, which involves identifying and delineating each object instance; or panoptic, which merges these two approaches for a comprehensive scene parsing. 
A selection of datasets that cater to these varied segmentation challenges is methodically enumerated in the second section of Table~\ref{tab:discriminative-datasets}, serving as a resource for models specializing in this task.

\subsubsection{Object Detection Datasets}
Object detection represents a hybrid task that fuses elements of localization and classification. It requires models to both identify the presence and ascertain the boundaries of objects within an image. 
The object detection datasets, which form a critical benchmark for models aiming to excel in this dual function, are curated towards the end of Table~\ref{tab:discriminative-datasets}. 
These datasets not only measure the model's accuracy in object identification but also its precision in bounding box prediction, thereby providing a comprehensive assessment of the model's object detection capabilities.

\begin{table*}[!h]
\centering
\caption{Datasets for Discriminative Visual Foundation Models}
\label{tab:discriminative-datasets}
\resizebox{1\textwidth}{!}
{
        \begin{tabular}{
        p{0.18\textwidth}
        p{0.23\textwidth}
        p{0.1\textwidth}
        p{0.1\textwidth}
        p{0.1\textwidth}
        p{0.33\textwidth}
        p{0.33\textwidth}
        p{0.2\textwidth}
        }
        \toprule
        \textbf{Task} & \textbf{Dataset} & \textbf{Year} & \textbf{Classes} &\textbf{Training} &\textbf{Testing} &\textbf{Evaluation Metric}& \textbf{Usage Examples} \\ 
        \midrule
        Image Classification & ImageNet-1k~\cite{ImageNet} & 2009 & 1000 & 1,281,167 & 50,000 & accuracy & model evaluation\\  
        &Food-101~\cite{Food-101} & 2014 & 102 & 75,750 &25,250 &accuracy & model evaluation\\
        &CIFAR-10~\cite{CIFAR} & 2009& 10 &50,000 &10,000 &accuracy & model evaluation\\
        & CIFAR-100~\cite{CIFAR} & 2009& 100 &50,000 &10,000 & accuracy & model evaluation\\
        & SUN397~\cite{SUN397} &2010 & 397 &19,850 &19,850 &accuracy& model evaluation\\
        & Stanford Cars~\cite{Stanford_Cars} &2013 & 196 &8,144 &8,041 &accuracy& model evaluation\\
        &FGVC Aircraft~\cite{FGVC_Aircraft} &2013 &100 &6,667 &3,333 &mean-per-class & model evaluation\\
        &Pascal VOC 2007 Classification~\cite{Pascal_VOC} &2007& 20 &5,011 &4,952 &11-point mAP & model evaluation\\
        &Describable Textures~\cite{textures} &2014& 47 &3,760 &1,880 &accuracy & model evaluation\\
        &Oxford-IIIT Pets~\cite{cats_and_dogs} &2012&37 &3,680 &3,669 &mean-per-class & model evaluation\\ &Caltech-101~\cite{Caltech-101} &2004& 102 & 3,060 &6,085 &mean-per-class & model evaluation\\
        &Oxford Flowers 102~\cite{Oxford-120-Flowers} &2008& 102 & 2,040 & 6,149 & mean per class & model evaluation\\
        \bottomrule
        \textbf{Task} &\textbf{Dataset} &\textbf{Year} &\textbf{Train}&\textbf{Val}& \textbf{Segmentation Type} & \textbf{Evaluation Metrics} & \textbf{Usage Example}\\
        \midrule
        Segment & Pascal VOC~\cite{Pascal_VOC} &2010&1464&1449& Semantic  & mIoU & model evaluation\\
        & COCO~\cite{MS-COCO} & 2014 &118k & 5k &Semantic/Instance/Panoptic& mIoU/mAP/PQ & model evaluation\\
        & Pascal Context~\cite{Pascal_context} &2014&4998 & 5105 &Semantic & mIoU& model evaluation\\
                & ADE20k~\cite{zhou2018semantic}& 2018 &20,210 & 2,000 &Semantic/Instance/Panoptic & mIoU/mAP/PQ & model evaluation\\
        & Cityscapes~\cite{cordts2016cityscapes} &2016 & 2,975 & 500 & Semantic/Instance/Panoptic & mIoU/mAP/PQ & model evaluation\\
        & PPDLS~\cite{MINERVINI201680} &2015 &810 &-- &Semantic &Symmetric Best Dice(SBD) &model evaluation/CVPPP Leaf segmentation challenge \\
        & BBBC038v1~\cite{BBBC038v1}&2019 &670 & 171 & Semantic/Instance & mIoU/mAP/PQ & model evaluation\\
        & TimberSeg~\cite{TimberSeg} &2022 &220 & -- & Semantic/Instance & mIoU/mAP/AR/AP50/F1-score & model evaluation\\
        & NDD20~\cite{NDD20} &2020 &3521 &881 & Semantic/Instance &mIoU/mAP &model evaluation\\
        & STREETS~\cite{STREETS} &2019 &2477 &523 &Semantic/Instance &mIoU/MAE &model evaluation \\
        & TrashCan~\cite{TrashCan} &2020 &7212 &-- &Semantic/Instance &AP &model evaluation \\
        \hline
        \textbf{Task} & \textbf{Dataset} & \textbf{Year} & \textbf{images} &\textbf{Boxes} &\textbf{Categories} &\textbf{Evaluation Metric}& \textbf{Usage examples}\\
        \midrule
        Object detection & Object365~\cite{Object365} &2019 &  638k & 10,101k & 365 & Avg. AP & model training/evaluation \\
        & OpenImages~\cite{OpenImages} &2020 & 1,515k &14,815k & 600 & Avg. AP & model training/evaluation \\
        & COCO~\cite{MS-COCO} & 2014& 123k & 896k & 80 &  Avg. AP & model training/evaluation \\

        \bottomrule
        
        \end{tabular}%
    }
\end{table*}

\subsection{Evaluative Metrics}

\subsubsection{Metrics for Image Classification}

In image classification, the following metric is predominantly used:

\begin{itemize}
    \item \textbf{Classification Accuracy}: This is the most straightforward metric for classification tasks, representing the proportion of correct predictions out of the total number of cases evaluated. It provides a direct measure of a model's capability to correctly label images, compared to a verified ground truth.
\end{itemize}

\subsubsection{Metrics for Image Segmentation}

Image segmentation models are evaluated using various metrics to ensure comprehensive and robust performance analysis:

\begin{itemize}
    \item \textbf{Mean Intersection-Over-Union (mIoU)}: mIoU is a critical metric for gauging the segmentation accuracy. It computes the average IoU scores, which measure the overlap between the predicted segmentation masks and the ground truth across all categories~\cite{minaee2020image}. This metric is particularly informative when evaluating a model's response to segmentation prompts, providing a quantitative measure of its segmentation precision in scenarios where specific points within an image are targeted for segmentation~\cite{SAM}.

    \item \textbf{Human Evaluation}: This qualitative assessment involves a panel of human judges evaluating the segmentation outputs against real-world scenarios~\cite{SAM}. It is indispensable for understanding the model's alignment with human visual interpretation and is often used in conjunction with quantitative metrics for a holistic evaluation.

    \item \textbf{Average Precision (AP)}: Employed in instance segmentation, AP measures the precision of model predictions at different IoU thresholds. It evaluates the model's ability to distinguish between individual object instances, reflecting its precision in both identifying the objects and delineating their boundaries accurately.

    \item \textbf{Average Recall at 1000 Proposals (AR@1000)}: Used in object proposal generation tasks, AR@1000 calculates the model's recall over the top 1000 proposed object regions~\cite{SAM}, providing insights into the model's capability to generate potential object locations before classification.
\end{itemize}

\subsubsection{Metrics for Object Detection}

The performance in object detection is measured through:

\begin{itemize}
    \item \textbf{Average Precision (AP)}: AP for object detection is akin to that used in instance segmentation, focusing on the accuracy of bounding box predictions. It evaluates the match between predicted and ground-truth boxes, crucial for assessing the model's localization accuracy.

    \item \textbf{Mean Average Precision (mAP)}: mAP aggregates the AP across all classes and/or over various IoU thresholds, providing a single measure of overall detection performance. It is a comprehensive metric that reflects the model's effectiveness across the entire detection spectrum, from localization to classification of objects within an image.
\end{itemize}


\subsection{Applications}


\paragraph{Baidu's Wenxin}
Baidu Wenxin has developed an array of computer vision products\footnote{https://wenxin.baidu.com/wenxin/cv}. The VIMER-CAE~\cite{VIMER-CAE} model excels in semantic segmentation, object detection, and instance segmentation. Meanwhile, VIMER-UFO 2.0\footnote{https://github.com/PaddlePaddle/VIMER/tree/main/UFO} showcases the 'All in One' and 'One for All'~\cite{VIMER-UFO} paradigms. 'All in One' has been recognized as the industry's largest computer vision model, with 17 billion parameters, and achieves state-of-the-art results in over 20 computer vision tasks. 'One for All' introduces a novel supernetwork and training scheme that caters to a range of visual tasks and hardware configurations, addressing challenges associated with large model parameters and inference performance.

\paragraph{Tencent's Hunyuan} Tencent's Hunyuan foundation model series is designed to address a wide array of computational tasks, including computer vision, natural language processing, understanding, and generation, as well as text-to-video generation. The HunYuan\_tvr model~\cite{HunYuan_tvr}, part of the Hunyuan series, is dedicated to Text-Video Retrieval, setting new benchmarks on multiple key datasets such as MSR-VTT~\cite{xu2016msr}, MSVD~\cite{wu2017deep}, and LSMDC~\cite{anne2017localizing}.

\paragraph{SenseTime's SenseNova} SenseTime's ``SenseNova'' suite of foundational models offers diverse capabilities ranging from natural language processing to automated data annotation and custom model training\footnote{https://www.sensetime.com/en/news-detail/51166818?categoryId=1072}. SenseNova's large language models are complemented by innovative generative AI models that facilitate text-to-image creation, digital human generation, and complex object generation. The model suite has enabled significant advancements in SenseTime's business domains, such as smart auto technologies, where it has contributed to the mass production of advanced perception systems and multimodal autonomous driving systems.

\paragraph{Huawei's Pangu CV} 
The Huawei Pangu series, launched in 2021, encompasses a suite of large models including those for NLP, computer vision, multimodal tasks, and scientific computing. Pangu CV\footnote{https://www.huaweicloud.com/product/pangu/cv.html}, notable for its size, integrates both discriminative and generative abilities, showcasing exceptional performance in small-sample learning on ImageNet.


\begin{table*}[!h]
     \centering  
     \caption{Summary of discriminative visual foundation model.} \label{tab:discriminative_models}
     \resizebox{1\textwidth}{!}{
         \begin{tabular}{
         p{0.17\textwidth} 
         p{0.15\textwidth} 
         p{0.25\textwidth} 
         p{0.07\textwidth} 
         p{0.2\textwidth} 
         p{0.18\textwidth} 
         p{0.2\textwidth} 
         p{0.18\textwidth} 
         p{0.08\textwidth}}

         \toprule
        \textbf{Name} & \textbf{Year(of first known publication)} & \textbf{Application} & \textbf{Base Model} & \textbf{Category} &\textbf{Resource}& \textbf{Num. Params} & \textbf{Training Time} & \textbf{Github Link}\\ 
 	\midrule
        CLIP~\cite{CLIP} &02/2021&Image Classification& - - & textually prompted models&V100 GPUs &  text encoder: a 63M-parameter 12layer 512-wide model with 8 attention heads & The largest ResNet model, RN50x64 model (18 days to train on 592 V100 GPUs), the largest Vision Transformer took 12 days on 256 V100 GPUs. & \href{https://github.com/OpenAI/CLIP}{link}\\

       ALIGN~\cite{ALIGN} &11/2017 & Image-Text Matching and Retrieval, Visual Classification & - - &textually prompted models & 1024 Cloud TPUv3 cores & - - & - - & - -\\ 

        Florence~\cite{Florence} &11/2021 & Zero-shot Transfer in Classification, Linear Probe in Classification, ImageNet-1K Fine-tune Evaluation, Few-shot Cross-domain Classification, Image-Text Retrieval & CLIP &textually prompted models & 512 NVIDIA-A100 GPUs & 893M parameters & 10 days to train on 512 NVIDIA-A100 GPUs with 40GB memory per GPU & - - \\

        BLIP~\cite{BLIP} &01/2022& Image-Text Retrieval, Image Captioning& - - &textually prompted models  & two 16-GPU nodes& - - & - - & \href{https://github.com/salesforce/BLIP}{link} \\

        SegGPT~\cite{SegGPT} & 04/2023& Video object segmentation & - - &visually prompted models& - - & - - & - - & \href{https://github.com/baaivision/Painter}{link} \\

        SAM~\cite{SAM} & 04/2023 & instance segmentation, Semantic segmentation, Panoptic Segmentation& - -&visually prompted models&256 A100 GPUs&- - & 256 A100 GPUS for 68 hours.&\href{https://segment-anything.com}{link} \\
        SEEM~\cite{SEEM} & 04/2023 & segmentation & - - & visually prompted models & - - & - - & - - &\href{https://github.com/UX-Decoder/Segment-Everything-Everywhere-All-At-Once}{link} \\
        CLIPSeg~\cite{CLIPSeg} & 12/2021 & image segmentation & - - &visually prompted models & - - & - - & - - & \href{https://eckerlab.org/code/ clipseg}{link}\\
        UniDetector~\cite{unidetector} & 03/2023 & object detection & - - &textually prompted models & - - & - - & - - &\href{https://github.com/zhenyuw16/UniDetector}{link}\\
       
     \bottomrule

     \end{tabular}
 }
 \end{table*}

\section{Multi-Modal Visual Foundation Models}

\subsection{Overview and Evolution}
\textit{Multi-modal Visual Foundation Models} (MVFM) represent a dynamic and rapidly evolving field within artificial intelligence, characterized by their ability to process and synthesize information from diverse data sources, including but not restricted to visual and textual inputs. 
Unlike the GVFM and DVFM that excel in single-modal tasks with fixed input-output patterns, MVFMs thrive on their proficiency in handling complex, multi-stage interactions among multiple modalities such as text, images, videos, audio, and more. 
These models pave the way for a myriad of applications, transcending traditional content generation and comprehension to embrace novel areas like cross-modal recommendation systems and interactive media.

\subsection{Foundational Concepts and Challenges}
The essence of MVFMs lies in their versatile architecture, designed to integrate and interpret a multiplicity of input modalities and deliver diverse output forms. 
This multimodality extends beyond visual and textual interplay, encompassing auditory, thermal, depth, and even data from inertial measurement units, thus offering a comprehensive sensory understanding. The training and fine-tuning of such models pose unique challenges, primarily due to the complexity and heterogeneity of multi-modal data, necessitating substantial, well-annotated datasets to achieve the desired learning outcomes.

\subsection{Network Architectures for MVFMs}

\paragraph{Model Architectures}
The domain of multi-modal learning has seen significant advancements with the introduction of various network architectures, each uniquely contributing to the field:
\begin{itemize}
    \item \textbf{MVP (Multimodality-guided Visual Pre-training)}~\cite{wei2022mvp} enhances the performance of Masked Image Modeling (MIM) by integrating multimodal semantic knowledge. Leveraging the CLIP vision branch, MVP excels in transferring knowledge to downstream tasks, demonstrating the power of multimodal guidance in visual pre-training.
    
    \item \textbf{M3 (Multimodal Model for Memory and Reasoning)}~\cite{ni2021m3p} offers a comprehensive approach to complex reasoning across text, images, and knowledge bases. By fusing vision, language, and memory, M3 is adept at tackling intricate multi-modal reasoning challenges.
    
    \item \textbf{VQ-VAE-2}~\cite{razavi2019generating} stands out for its generative capabilities in both the visual and auditory realms, addressing tasks that span across image and audio data domains.
\end{itemize}

\paragraph{Loss Function}
Crafting loss functions for multi-modal learning is a delicate process, influencing the efficacy of the model in handling diverse modalities.
\begin{itemize}
    \item \textbf{Contrastive Loss} is a cornerstone in models like CLIP~\cite{radford2021learning}, facilitating the learning of a joint embedding space where similar items are brought closer together, promoting coherence across modalities.\cite{wang2021understanding}

    \item \textbf{Modality Matching Loss} ensures alignment of representations across modalities, fostering a shared understanding and enhancing the model's ability to integrate multi-modal information. \cite{chen2022maestro}

    \item \textbf{Reconstruction Loss} is pivotal in generative tasks, enabling models to accurately replicate images or audio, thus maintaining the integrity of the generated modality. \cite{wang2018reconstruction}

    \item \textbf{Joint Loss} amalgamates modality-specific losses, propelling the model to excel across all considered modalities.\cite{li2017person} 

    \item \textbf{Custom Loss Functions} are often crafted to address specific multi-modal learning challenges, tailored to the requirements of the task at hand.\cite{barton2021convolution} 
\end{itemize}

\subsection{Datasets}
The foundation of robust MVFMs is the quality and scale of the datasets on which they are trained. Here we present a concise overview of benchmark datasets facilitating the training of multi-modal models:
\begin{itemize}
    \item \textbf{COCO Dataset}~\cite{chen2015microsoft}: A cornerstone in multi-modal research, COCO offers over 200,000 images annotated with object segments, categories, and human key points. It is enhanced with descriptive captions, making it invaluable for both object-centric tasks and natural language-based image processing.

    \item \textbf{Flickr Image-Caption Datasets}~\cite{mao2014explain, young2014image}: The Flickr8k and Flickr30k datasets are repositories of images paired with multiple descriptive captions, with Flickr30k offering a more extensive collection. They serve as a platform for image captioning and related language-visual tasks.

    \item \textbf{Conceptual Captions}~\cite{sharma2018conceptual}: This dataset boasts over 3.3 million image-text pairs, offering a breadth of human-generated descriptions that span various subjects, thereby providing a rich resource for training models on a wide spectrum of visual-textual contexts.

    \item \textbf{M5Product}~\cite{dong2022m5product}: A comprehensive dataset featuring five modalities, including image, text, table, video, and audio. It is one of the largest of its kind, accommodating thousands of categories and attributes, which is pivotal for training models to comprehend and generate content across multiple modalities.
\end{itemize}
MVFMs trained on such datasets have demonstrated remarkable proficiency in numerous multi-modal tasks, including but not limited to image captioning, visual question answering, and audio-visual scene understanding. 
The continuous enrichment of these datasets is crucial for the development of even more capable and generalizable multi-modal models, paving the way for groundbreaking applications across diverse AI domains.

\subsection{Challenges in MVFM Development}
Developing multi-modal models entails overcoming several intricate challenges:

\begin{itemize}
    \item \textbf{Data Collection and Labeling} requires a more complex and costly process compared to single-modal data due to the need for comprehensive multi-modal datasets, often compounded by data noise and alignment issues.

    \item \textbf{Model Architecture} design is a sophisticated task, demanding strategies to effectively process and integrate information from disparate modalities while preserving their unique properties.

    \item \textbf{Pre-Training Objectives} must be innovatively designed to support unsupervised learning across modalities, with objectives tailored to the nuances of multi-modal data.

    \item \textbf{Computational Demands} for training multi-modal models are significant, posing challenges in scaling and managing computational resources.

    \item \textbf{Technique Migration} from small-scale to large-scale models necessitates exploration and adaptation to ensure the efficacy of techniques at a larger scale.

    \item \textbf{Cross-Modal Coupling and Decoupling} problems require careful consideration within the model framework to establish dynamic modality relationships effectively.
\end{itemize}


\subsection{Representative Examples of MVFMs}
MVFMs represent a paradigm shift in artificial intelligence, offering an integrated approach to understanding and generating multi-faceted content. 

\paragraph{Visual ChatGPT}
Inspired by the linguistic prowess of ChatGPT, Visual ChatGPT~\cite{Visual_ChatGPT} extends its capabilities into the visual domain. 
This enhanced system integrates an array of VFMs and is steered by a sophisticated Prompt Manager. This component is vital for three reasons: it informs ChatGPT about the functions and data formats of each VFM, it translates visual data such as images or depth maps into a language format for ChatGPT's comprehension, and it adeptly manages the historical context, priority, and potential conflicts among VFMs. 
The iterative process of feedback and adjustment allows Visual ChatGPT to refine responses, providing a comprehensive multi-modal dialogue experience that caters to both textual and visual queries.

\paragraph{PandaGPT}
PandaGPT~\cite{su2023pandagpt} is designed as a versatile instruction-following model with sensory abilities akin to human sight and hearing. 
Its proficiency lies in executing complex tasks that range from crafting elaborate image narratives to creating stories inspired by videos and responding to audio-based inquiries. 
This model exemplifies the potential of multi-modal AI in engaging with and interpreting a world that is as visual and auditory as it is textual.

\paragraph{NExT-GPT}
NExT-GPT~\cite{wu2023next} represents a significant leap in multi-modal AI, serving as a comprehensive any-to-any model adept at input and output across texts, images, videos, and audio. Its architecture is tripartite: it encodes diverse modalities, harnesses the reasoning capabilities of a LLM, and generates outputs in the chosen modalities. 
The model overcomes the constraints of previous multi-modal LLMs by delivering multi-faceted content generation. Its innovations are manifold:
\begin{enumerate}
\item Universal multi-modal comprehension is achieved through the integration of multimodal adaptors and diffusion decoders with an LLM.
\item Versatility in content generation is provided, allowing for input and output across various modalities.
\item Efficiency is enhanced by employing pre-trained encoders and decoders, minimizing training needs and expanding the model's adaptability.
\item Advanced cross-modal understanding and content generation are supported through modality-switching instruction tuning (MosIT) and a curated dataset.
\item The development of a unified AI agent is exemplified, showcasing the potential to emulate human-like modality processing and interaction.
\end{enumerate}

\section{Future Directions in Visual Foundation Models}

\subsection{Evolving Landscape of Discriminative Models}
The advancement of foundation models in discriminative tasks is unfolding rapidly, with the aim of unifying various levels of task granularity. 
Researchers are delving into the integration of tasks ranging from image-level classifications to pixel-level segmentations. 
Although models like CLIP~\cite{CLIP} and ALIGN~\cite{ALIGN} have shown adaptability across diverse tasks, the quest for a truly universal vision interface that mirrors the versatility of GPT in language tasks is ongoing. 
The potential for creating interactive and prompt-responsive models, similar to ChatGPT, is being explored by innovative models like SAM~\cite{SAM} and SEEM~\cite{SEEM}. 
These models show promise in enabling a broader range of tasks through zero-shot learning and other flexible approaches, yet there is still significant potential to expand their task coverage and interactivity.

\subsection{Convergence of Generative and Discriminative Capabilities}
\begin{figure}[h!]
 \centering
\includegraphics[width=0.9\linewidth]
{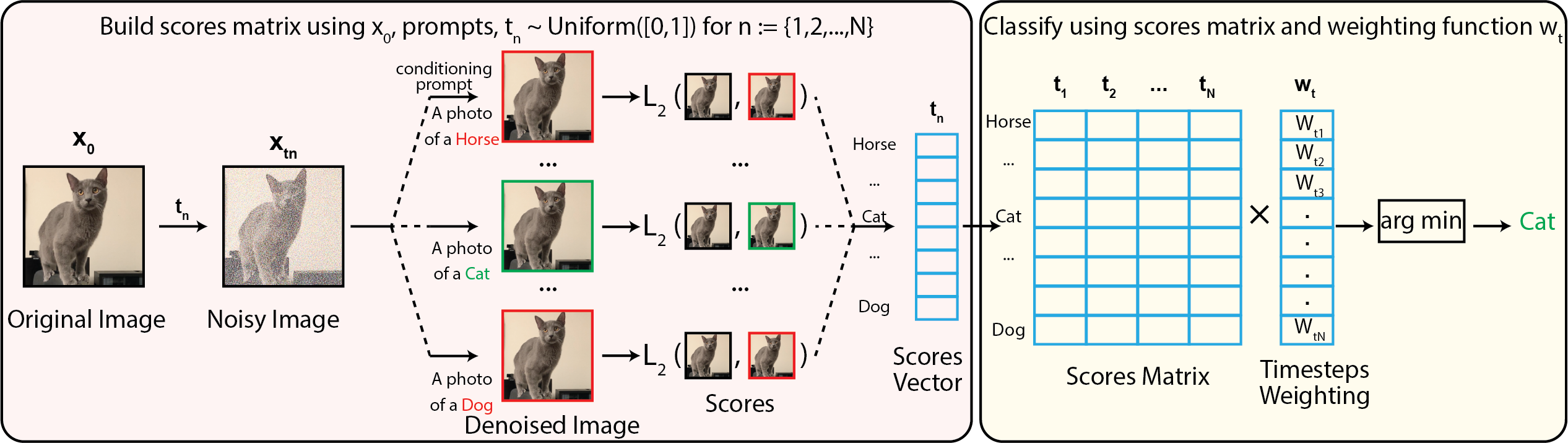}
\caption{Schematic representation of employing diffusion models for zero-shot classification tasks. This process exemplifies the model's proficiency in synthesizing and refining visual data to align with textual prompts, culminating in accurate category identification.}
\label{fig:Zero-Shot Classification using Diffusion Models}
\end{figure}

A unified foundation model that amalgamates generative and discriminative functions is an aspiration in the field, with the current landscape featuring models dedicated to either function. 
While generative models like Stable Diffusion~\cite{stable_diffusion} excel in creating images from text prompts, their application in discriminative tasks is an area of active research. 
Efforts are underway to leverage generative models' representational learning for tasks such as segmentation and object detection~\cite{diffdis,ge2023generation,gu2022diffusioninst,chen2023diffusiondet,ni2023refdiff,segdiff,li2023diffusion}. The example in Figure~\ref{fig:Zero-Shot Classification using Diffusion Models} illustrates the potential of using diffusion models for classification tasks.

Generative models have shown resilience in inference and generalization across various datasets for discriminative tasks. 
However, their computational intensity poses a challenge for practical applications. 
Research focused on reducing computational costs while retaining robust performance is a promising direction.
DiffDis~\cite{diffdis} is a notable step towards unifying generative and discriminative training within the diffusion framework. 
By conditioning text embeddings on input images, DiffDis utilizes the same network for both tasks, demonstrating impressive zero-shot classification and retrieval capabilities. 
Expanding this approach to other discriminative tasks could lead to more integrated and efficient models.

\subsection{Integrating Diverse Modalities}
The ability to process and generate content across multiple modalities is a frontier yet to be fully conquered. 
While models like NExT-GPT~\cite{wu2023next} have expanded their repertoire to include language, images, videos, and audio, the integration of other modalities, such as 3D vision, is ripe for exploration. 
Enhancing the quality and richness of generative outputs remains an important objective, with significant room for improvement in generating high-fidelity multimedia content.


\section{Conclusion}
This review paper has charted the evolution and current state of Visual Foundation Models, underscoring their transformative impact on computer vision and related fields. We have dissected the progression of GVFM, spotlighting the strides made by industry-leading products that leverage these models. Furthermore, our exploration of DVFM provided a detailed taxonomy and an examination of their applications, including but not limited to image classification, object detection, and segmentation.

The intersection of GVFM and DVFM has emerged as a fertile ground for innovation, revealing the potential of generative models in discriminative tasks. Our analysis extended to the synergy between VFMs and LLMs, pinpointing areas for growth in model interactivity and user engagement. This synthesis of ideas aims to propel the field towards a new frontier where VFMs not only exhibit advanced visual understanding but also interact seamlessly across a multitude of tasks.

As the landscape of VFMs continues to evolve, we anticipate further amalgamation of generative and discriminative capabilities, leading to more holistic and sophisticated models. The drive towards creating interactive, multi-modal, and adaptive systems represents the next leap forward, promising to unlock unprecedented applications and capabilities. It is our hope that this review will act as a catalyst for continued innovation and research, shaping the future trajectory of computer vision technology.
\bibliography{sn-bibliography} 
\end{document}